\documentclass[acmsmall]{acmart}

\renewcommand\footnotetextcopyrightpermission[1]{} 
\usepackage[utf8]{inputenc}
\usepackage{amsmath}
\usepackage{hyperref}
\usepackage{amsfonts}
\usepackage{graphicx}
\usepackage{algorithm}
\usepackage{algpseudocode}
\usepackage{float}
\usepackage{booktabs}
\usepackage{mathtools}
\usepackage{etoolbox}

\usepackage{caption}

\usepackage{subcaption}

\usepackage{siunitx}

\newcount\Comments  
\usepackage{color}
\definecolor{darkgreen}{rgb}{0,0.5,0}
\definecolor{purple}{rgb}{1,0,1}
\newcommand{\kibitz}[2]{\ifnum\Comments=0\textcolor{#1}{#2}\fi}

\begin{document}

\title{Streaming data preprocessing via online tensor recovery for large environmental sensor networks}

\author{Yue Hu}
\affiliation{%
  \institution{Vanderbilt University}
  \streetaddress{1025 16th Ave S, Suite 102}
  \city{Nashville}
  \country{USA}
}
\email{yue.hu@vanderbilt.edu}

\author{Ao Qu}
\affiliation{%
  \institution{Vanderbilt University}
  \streetaddress{1025 16th Ave S, Suite 102}
  \city{Nashville}
  \country{USA}
}
\email{ao.qu@vanderbilt.edu}

\author{Yanbing Wang}
\affiliation{%
  \institution{Vanderbilt University}
  \streetaddress{1025 16th Ave S, Suite 102}
  \city{Nashville}
  \country{USA}
}
\email{yanbing.wang@vanderbilt.edu}

\author{Daniel B. Work}
\affiliation{%
  \institution{Vanderbilt University}
  \city{Nashville}
  \country{USA}
}
\email{dan.work@vanderbilt.edu}

\begin{abstract}
Measuring the built and natural environment at a fine-grained scale is now possible with low-cost urban environmental sensor networks. However, fine-grained city-scale data analysis is complicated by tedious data cleaning including removing outliers and imputing missing data. While many methods exist to automatically correct anomalies and impute missing entries, challenges still exist on data with large spatial-temporal scales and shifting patterns. To address these challenges, we propose an online robust tensor recovery (OLRTR) method to preprocess streaming high-dimensional urban environmental datasets. A small-sized dictionary that captures the underlying patterns of the data is computed and constantly updated with new data. OLRTR enables online recovery for large-scale sensor networks that provide continuous data streams, with a lower computational memory usage compared to offline batch counterparts. In addition, we formulate the objective function so that OLRTR can detect structured outliers, such as faulty readings over a long period of time. We validate OLRTR on a synthetically degraded National Oceanic and Atmospheric Administration temperature dataset, with a recovery error of 0.05, and apply it to the Array of Things city-scale sensor network in Chicago, IL, showing superior results compared with several established online and batch-based low rank decomposition methods.
\end{abstract}



\maketitle

\section{Introduction}
\subsection{Motivation}
The United Nations established 17 Sustainable Development Goals that are to be achieved by 2030~\cite{UNreport}. One of the goals is to promote sustainable and resilient, inclusive and safe cities. 
To quantify the effects of the built environment on micro climate and other environmental impacts, many urban-scale environmental sensing initiatives exploiting emerging 
\textit{internet of things} (IoT) technologies are being developed (e.g.,~\cite{Catlett:2017:ATS:3063386.3063771,citysense}). These projects measure block-by-block micro-climate quantities to inform better green infrastructure investment, transportation planning, and energy-saving designs. 

Despite the high spatial resolution information provided by the low-cost sensors, data quality and data treatment still remain major concerns that hinders a wider adoption of these technologies~\cite{lewis2018low,karagulian2019review}. Outliers and missing data are amongst the challenges. Current approaches to clean the datasets prior to interpretation are often limited in functionality for which anomalies or missing data are independently addressed~\cite{DASZYKOWSKI2007203,HILL20101014,RPCA_em}. 

\begin{figure}
    \centering
    \includegraphics[width=0.45 \linewidth]{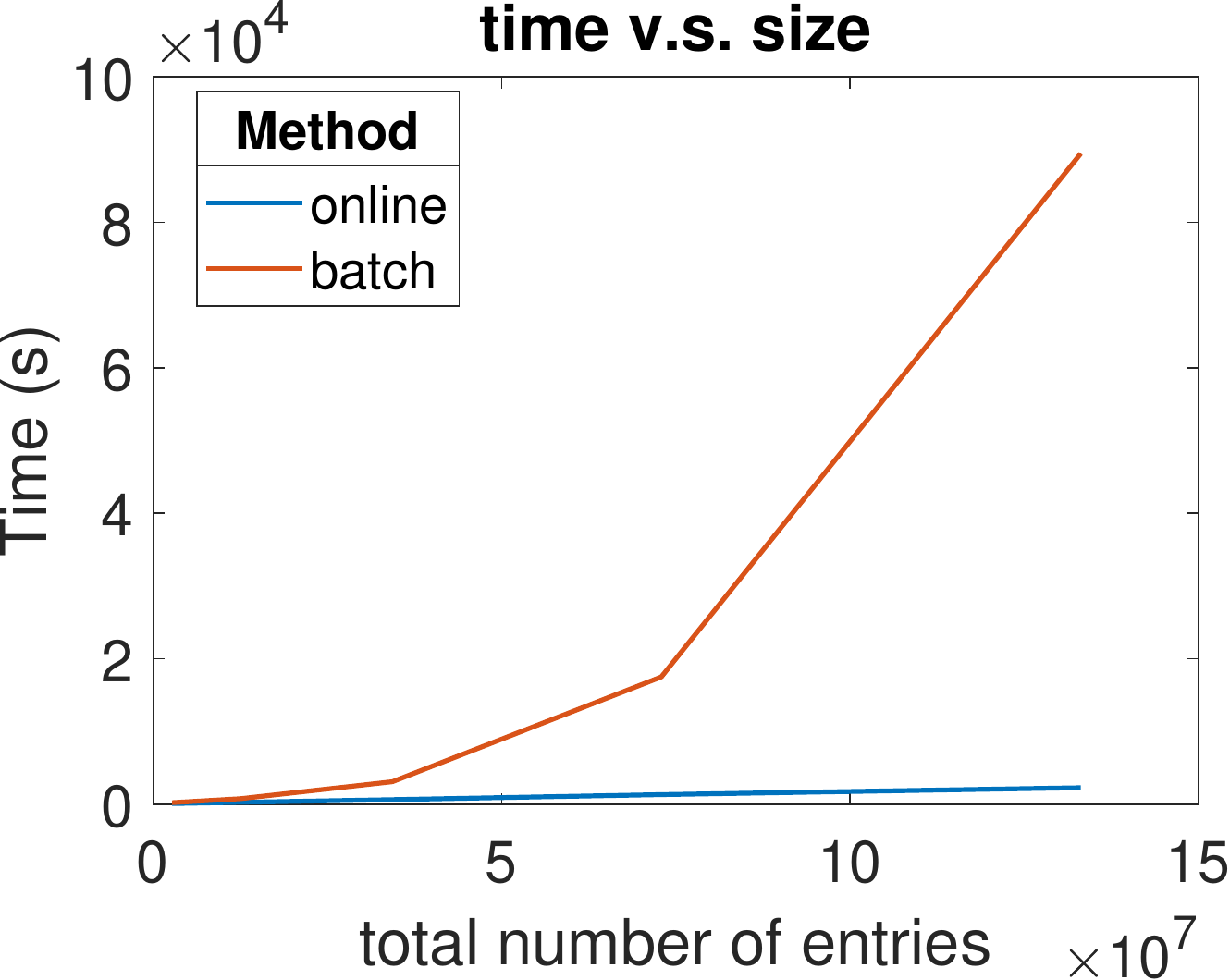}
    \caption{Computation time as a function of input tensor size for batch-based method~\cite{hu2020robust} and our online-based method (proposed). For batch-based tensor decomposition, computation time increases sharply with data size, thus impractical for large scale systems.}
    \label{fig:batch_OL}
\end{figure}

A promising direction to overcome these limitations involves tensor factorization methods, which have shown state-of-the-art performance in detecting outliers and imputing missing data~\cite{CHEN201966,hu2020robust}. Because large-scale urban sensor networks usually produce higher-order data that contains spatial and temporal relationships, low-rank recovery can be naturally applied. Yet these batch-based methods rely on collecting the data samples for all time, and re-solving the problem when new data arrives, making them sub-ideal for streaming sensor networks deployed for continuous monitoring.  As depicted in Fig.~\ref{fig:batch_OL}, computation time increases sharply with data size, posing challenge for the real-world applications.

\subsection{Contribution}
\textbf{To overcome the challenges of batch tensor factorization methods, we develop an \textit{online robust tensor recovery} (\textbf{OLRTR}) algorithm to pre-process streaming data from large-scale urban sensor networks.} 

The main contribution of this work is to introduce OLRTR to automatically correct errors and impute missing data common to large distributed urban sensor networks.
OLRTR computes and sequentially updates a small-sized dictionary that stores the underlying, time-varying patterns of the data, which significantly lowers the memory usage and can adapt to shifting patterns in the datasets. 

Two real-world experiments demonstrate the effectiveness of OLRTR. The first uses a  complete, high quality \textit{National Oceanic and Atmospheric Administration} (NOAA) temperature dataset~\cite{noaa_data}, which is artificially degraded by injecting known outliers and also by removing some entries to simulate missing data. We demonstrate that the proposed tensor factorization approach correctly identifies the outliers and recovers accurate values for the missing data. The second experiment applies the method to raw and incomplete temperature data from the state of the art IoT platform known as the \textit{Array of Things} (AoT) urban sensing platform in Chicago, IL~\cite{aot_data}. The recovered temperature data is validated by comparing to nearby NOAA readings. The experimental results show the superiority of OLRTR over several online and batch-based methods, as well as the potential of OLRTR to provide reliable data streams for real-world sensor networks.

\subsection{Overview of the proposed method}
We briefly summarize the batch-based tensor factorization approach to remove outliers and impute missing data, then give an overview how we adapt it to online settings. 

\begin{figure}
\centering
\includegraphics[width=\linewidth]{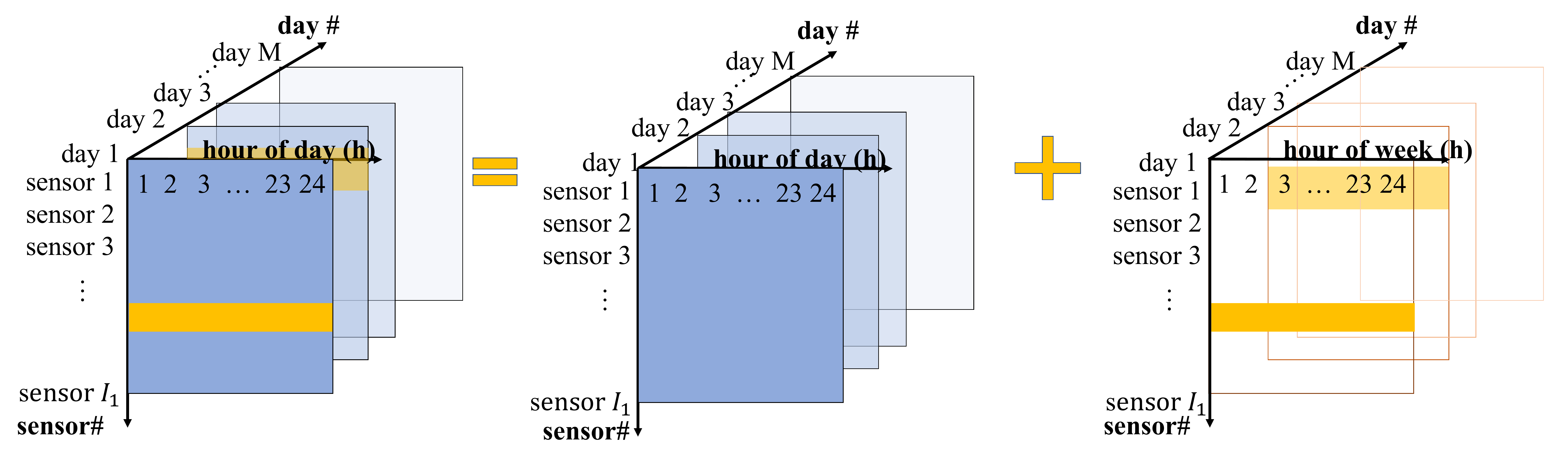}
\caption{Observation data decomposed into low rank tensor for clean data and fiber-sparse tensor for outlier data caused by malfunctioning sensors.}
\label{fig:decom_illu}
\end{figure}

The sensor observation data is organized in a tensor~\cite{kolda2009tensor,goldfarb2014robust}, to exploit the spatial and temporal structures in the data. An example of a three-way tensor storing sensor data is shown in the first column of Fig.~\ref{fig:decom_illu}, where the first mode corresponds to each sensor, the second mode to each hour in a 24-hour period, and the third mode to each 24-hour period in the dataset. Tensor factorization approaches~\cite{kolda2009tensor,goldfarb2014robust} exploit the fact that such large, noisy, and incomplete datasets actually have low intrinsic dimensionality. Furthermore we assume that the outliers in the sensor network have a specific sparsity pattern, persisting across one of the orders of the tensor, as shown in the third column of Fig.~\ref{fig:decom_illu}. This outlier structure reflects the observation that some sensors degrade and produce faulty data for extended periods of time. 

To reconstruct the underlying clean complete data and detect the outliers, we solve the following optimization problem for $M$ days of data: 
\begin{equation} 
\label{eq:intro}
\begin{aligned}
    & \underset{\mathcal{X}^M,\mathcal{E}^M}{\text{min}}
    & &\text{rank}(\mathcal{X}^M) + \lambda  \text{  sparsity} (\mathcal{E}^M)\\ 
    & \text{s.t.}
    & & \mathcal{B^M}_{i_1i_2\dots i_N}= (\mathcal{X}^M+ \mathcal{E}^M)_{i_1i_2\dots i_N}, \\
    & &&\text{where } (i_1,i_2,\dots,i_N) \text{ is an observed entry, }
\end{aligned}
\end{equation}
where tensor $\mathcal{X}^M$ represents the clean complete data, tensor $\mathcal{E}^M$ denotes the outliers, and $\mathcal{B}^M$ denotes the observation data. The size of $\mathcal{X}^M$, $\mathcal{E}^M$ and $\mathcal{B}^M$ grows with the number of days $M$. We regularize the low dimensionality of $\mathcal{X}^M$ measured by the Tucker rank~\cite{tucker1966some}, and the sparsity of the outlier tensor $\mathcal{E}^M$, under the constraint that $\mathcal{X}^M$ and $\mathcal{E}^M$ adds up to the raw data $\mathcal{B}^M$ in the observed entries.

In the batch-based approach~\cite{hu2020robust}, Problem~\eqref{eq:intro} is solved by singular value thresholding~\cite{cai2010singular,candes2011robust} based on the \textit{alternating direction method of multipliers} (ADMM) framework~\cite{hu2020robust,goldfarb2014robust} in an iterative manner. However, singular value thresholding can only be computed after all samples are collected. This practically means that after we conduct the tensor decomposition at day $M$, when new data comes in on day $M+1$, we have to solve Problem~\eqref{eq:intro} from scratch. It is inefficient that we cannot reuse the results from earlier computations, and that we have to store all  observation data in memory, which can grow large quickly. Moreover, in each iteration of solving~\eqref{eq:intro}, we need to compute a singular value decomposition, which is computationally expensive especially as the tensor size grows. Thus, the batch-based method is hardly scalable to large streaming systems.

To enable online processing and compute the data in a sequential manner, we develop OLRTR, which regularizes the rank of tensor $\mathcal{X}$ in a new way. Namely, we keep a small dictionary forming a basis for the low-rank subspace, and find the corresponding coefficients to represent $\mathcal{X}$ in terms of the basis. The size of the dictionary gives an upper bound on the rank of $\mathcal{X}$, and we aim to find the dictionary to best capture all the samples. In practice, we update the dictionary after each sample estimation so that the dictionary can also adapt to the shifting dynamics of the underlying subspace. In this way, the computation for each sample is decoupled. Furthermore, we only need a small space to store the dictionary, and there is no need to store all observation data in memory. Thus, our approach enables online recovery for large-scale sensor networks. 

The remainder of this article is as follows. Section~\ref{Sec:related} reviews the most related works. Section~\ref{Sec:Pre} introduces the basic tensor notations, and reviews batch-based tensor robust decomposition. Section~\ref{sec:Method} develops our proposed OLRTR method. Section~\ref{sec:experiment} shows our experiments on both synthetic and real-world datasets. We finally conclude the work in Section~\ref{Sec:conclude}. 

\section{Related work} \label{Sec:related}
\subsection{Other data preprocessing efforts}
Data preprocessing is fundamental to building a reliable and comprehensive understanding of the analysis tasks afterwards. In general, data preprocessing techniques can be categorized into three aspects~\cite{famili1997data}: (1) \textit{data transformation}, such as data filtering and noise modeling, (2) \textit{information gathering} such as visualization and feature extraction, and (3) \textit{generation of new information} such as time series analysis, data fusion and simulation/creation of new features. These treatments help to solve the problems that hinders further analysis and provide us with meaningful understanding of the measurements.

\subsection{Batch-based low-rank learning}
Low rank learning has been widely used to exploit the correlations in the datasets, with application in video surveillance~\cite{candes2011robust}, link prediction~\cite{dunlavy2011temporal}, anomaly detection~\cite{li2018multi}, and so on.

Two threads of works are most related with ours, namely robust matrix and tensors decomposition with gross corruption, and low rank matrix and tensor completion with missing data. For robust decomposition, $l_1$ norm  is usually used as an convex regularization for element-wise sparsity~\cite{goldfarb2014robust}. Cauchy distribution and the chi-squared distribution is also used to deal with the case when gross corruption and small noises co-exists~\cite{wu2017robust, yang2015robust}. If the outlier is structured, for example grouped in columns, then $l_{2,1}$ norm  regularization is usually used~\cite{xu2010robust,zhou2017outlier,hu2020robust}. Regarding missing data imputation in tensors, CANDECOMP/PARAFAC (CP) decomposition~\cite{hitchcock1927expression} is used in the works~\cite{zhao2015bayesian,wu2018fused}, and Tucker decomposition~\cite{tucker1966some} is used in the works~\cite{chen2013simultaneous, hu2020robust}.

However, the above mentioned methods are all batch-based, requiring all samples be collected before the low rank decomposition can be performed. This does not meet our need in sensor networks where we need the estimation to be performed continuously as new data comes in. Moreover, the memory and computation requirement of batch methods poses challenge for large-scale sensor networks. 

\subsection{Online low-rank learning}
To adapt the low-rank learning to online settings, various attempts in matrix and tensor fields have been made. For matrix decomposition, Mairal et al.~\cite{mairal2010online} develops an online dictionary learning method based on the assumption of sparse coding, i.e. the data vectors are linear combinations of a few of the basis vectors. Inspired by~\cite{mairal2010online}, Feng et al.~\cite{feng2013online} proposes matrix online robust PCA via stochastic optimization, which is provably robust to sparse corruption. Shen et al.~\cite{shen2016online} develops a \textit{Low-Rank Representation} (LRR) based online algorithm that segments data generated from a union of subspaces with improved time complexity and memory footprint. He et al.~\cite{he2011online} tracks the subspaces by gradient descent on Grassmannian, the manifold of all $d$-dimensional subspaces. For tensor decomposition, Sobral et al.~\cite{sobral2015online} proposes an \textit{online stochastic framework for tensor decomposition} (OSTD) for video sequence background subtraction. Li et al.~\cite{li2019online} develops an \textit{online robust low-rank tensor modeling} (ORLTM) method that can deal with streaming tensor data drawn from a mixture of multiple subspaces effectively through dictionary learning. Our work is most inspired by~\cite{feng2013online}, but we extend the decomposition from matrix to tensor to exploit the multi-dimensional correlations, and we also adapt to the scenario of structured outlier and missing data.

There are two major differences between our method and the methods mentioned above. First, we aim to find structured outliers grouped in tensor fibers, while the approaches~\cite{li2019online,sobral2015online,feng2013online,shen2016online,he2011online} only deals with unstructured element-wise outliers. While there are online decomposition approaches to find column outliers~\cite{mateos2012robust}, they deal with the case when an entire sample of data vector is an outlier. In comparison, in our case the structured outliers lies across multiple samples. For example, in sensor networks we aim at finding out malfunctioning sensors producing wrong recordings for a consecutive time, while approaches~\cite{mateos2012robust} can only find out abnormal time slots when all sensors deviate from normal recording. We solve this problem by conducting decomposition on minibatches instead of single samples. Second, we are able to deal with the case when outliers and missing data co-exists, whereas the approaches~\cite{li2019online,sobral2015online,feng2013online,shen2016online} only considers outliers with full observations. Several online robust decomposition approaches handle missing data, including~\cite{he2011online} via gradient descent on Grassmannian, and Kasai et al.~\cite{kasai2016network} based on  the Candecomp/PARAFAC (CP) decomposition, yet they both deal with element-wise outliers.

\section{Preliminaries}
\label{Sec:Pre}

\subsection{Tensor basics}
We briefly introduce our notation and define tensor operators, following a standard notation~\cite{kolda2009tensor, goldfarb2014robust, hu2020robust}, (See also ~\cite{kolda2009tensor,goldfarb2014robust} for a detailed discussion).

A tensor is denoted by an Euler script letter (e.g., $\mathcal{X}$); and a matrix by a boldface capital letter (e.g., $\mathbf{X}$); a vector by a boldface lowercase letter (e.g., $\mathbf{x}$); and a scalar by a lowercase letter (e.g., $x$). A tensor of order $N$ has $N$ dimensions. 
A \textit{fiber} is a column vector formed by fixing all indices of a tensor but one.

The \textit{unfolding} function flattens the tensor into a matrix to facilitate the computation. The unfolding of a tensor $\mathcal{X} \in \mathbb{R}^{I_1 \times I_2 \times \dots \times I_N}$ in the $n^{\text{th}}$ mode is formed by rearranging the mode-$n$ fibers as its columns, resulting in a matrix $\mathbf{X}_{(n)} \in \mathbb{R}^{I_n \times I_{N \setminus n}}$, where $I_{N \setminus n} = \prod_{i \neq n, i=1}^N I_i = I_1 \times \dots \times I_{i-1} \times I_{i+1} \times \dots \times I_N$. 
To convert the unfolding matrix back to original tensor, the \textit{fold} function is applied: $\text{fold}_n(\mathbf{X}_{(n)} )= \mathcal{X}.$
    
The inner product of $\mathcal{X,Y} \in \mathbb{R}^{I_1 \times I_2 \times \dots \times I_N}$ is the sum of their element-wise product: 
$ \langle \mathcal{X,Y} \rangle = \sum_{i_1=1}^{I_1}\sum_{i_2=1}^{I_2} \dots \sum_{i_N=1}^{I_N}x_{i_1i_2\dots i_N}y_{i_1i_2\dots i_N},$
where $x_{i_1i_2\dots i_N}$ and $y_{i_1i_2\dots i_N}$ denote the $(i_1,i_2, \cdots, i_N)$ element of $\mathcal{X}$ and $\mathcal{Y}$ respectively. 

The \textit{tensor Frobenius norm} follows naturally from the matrix Frobenius norm, and is defined as:
$\|\mathcal{X}\|_F = \sqrt{\langle\mathcal{X},\mathcal{X}\rangle}.$

The mode-$n$ product of a tensor $\mathcal{X}\in \mathbb{R}^{I_1 \times I_2 \times\dots \times I_N}$ and a matrix $\mathbf{A} \in \mathbb{R}^{J \times I_n}$ is denoted by $\mathcal{X} \times_n \mathbf{A} = \mathcal{Y}$, where $\mathcal{Y} \in \mathbb{R}^{I_1 \times I_2 \times \dots  \times I_{n-1} \times J \times I_{n+1} \times\dots \times I_N}$.

The \textit{Tucker decomposition}~\cite{goldfarb2014robust,kolda2009tensor} approximates a tensor  $\mathcal{X} \in \mathbb{R}^{I_1 \times I_2 \times\dots \times I_N}$ as a core tensor $\mathcal{G} \in \mathbb{R}^{c_1 \times c_2 \times\dots \times c_N}$ multiplied in each mode $n$ by an appropriately sized matrix $\mathbf{U}^{(n)}$: $\mathcal{X} \approx \mathcal{G} \times_1 \mathbf{U}^{(1)}\times_2 \mathbf{U}^{(2)}\times\dots \times_N \mathbf{U}^{(N)}.$
The matrices $\mathbf{U}^{(n)} \in \mathbb{R} ^{I_n \times c_n}$ are factor matrices, which are usually assumed to be orthogonal. 

\subsection{Robust tensor decomposition}
In this section, we briefly summarize the batch-based higher-order tensor decomposition with fiber-wise corruption as posed in~\cite{hu2020robust}, which we adapt to online settings in Section~\ref{sec:Method}.

The batch-based setup with complete observation is as follows. We are given a high dimensional data tensor $ \mathcal{B} $ that is corrupted in a few fibers. In other words, we have $\mathcal{B = X +E}$, where  $\mathcal{X}$ is the low rank tensor, and $\mathcal{E}$ is the sparse fiber outlier tensor. $\mathcal{B, X, E}  \in \mathbb{R}^{I_1 \times I_2 \times\dots \times I_N}$.
Our goal is to reconstruct $\mathcal{X}$ on the non-corrupted fibers, as well as identify the outlier location. Without loss of generality, we assume the fiber-wise corruption occurs along the first mode. The optimization problem goes as follows:

\begin{equation}
\label{eq:TPCA_batch}
\begin{aligned}
& \underset{\mathcal{X}_i,\mathcal{E}}{\text{min}}
& & \sum_{i=1}^N\|\mathbf{X}_\mathnormal{i(i)}\|_*+ \lambda  \|\mathbf{E}_{(1)}\|_{2,1} \\
& \text{s.t.}
& & \mathcal{B = X_\mathnormal{i}+E}, i = 1,2,\dots,\ N,
\end{aligned}
\end{equation}
where  $\mathcal{X}_1, \mathcal{X}_2, \dots, \mathcal{X}_N \in \mathbb{R}^{I_1 \times I_2 \times\dots \times I_N}$ are auxiliary variables split from the same variable $\mathcal{X}$ to decouple the computation in different tensor modes, and $\mathbf{X}_\mathnormal{i(i)}$ is the corresponding mode-$i$ unfolding for $\mathcal{X}_i$. $\mathbf{E}_{1}$ is the unfolding of $\mathcal{E}$ in the fiirst mode. The $N$ constraints $\mathcal{B = X_\mathnormal{i}+E}$ ensure that $\mathcal{X}_1, \mathcal{X}_2, \dots, \mathcal{X}_N$ are all equal to the original low rank tensor $\mathcal{X}$. The sum of nuclear norms $\sum_i\|\mathbf{X}_{i(i)}\|_*$ is a convex relaxation for Tucker rank of $\mathcal{X}$~\cite{goldfarb2014robust}, with the nuclear norm computed as $\|\mathbf{X}\|_* := \sum_i\sigma_i$, where $\sigma_i$ denotes the $i$-th singular value of $\mathbf{X}$.  The $l_{2,1}$ norm of a matrix $\mathbf{E} \in \mathbb{R}^{I_1 \times I_2}$ is used to encourage the column-wise sparsity, defined as $ \|\mathbf{E}\|_{2,1} = \sum_{j=1}^{I_2}\sqrt{\sum_{i=1}^{I_1}(e_{ij})^2}.$

We note that the mode along which to unfold $\mathcal{E}$ in~\eqref{eq:TPCA_batch} depends on what kind of outliers we want to detect. For example, in sensor network data as shown in Fig.~\ref{fig:decom_illu}, unfolding $\mathcal{E}$ along first mode corresponds to abnormal hours when records from all sensors deviates from normal; second mode corresponds to abnormal sensors that records wrong value for a consecutive period-- which is our goal in this paper. To simplify the notations, the rest of the paper is based on unfolding along the first mode, but the method for other modes follows the same line. 

In addition to observation data being grossly corrupted, we might have only partial observations of $\mathcal{B}$, and we seek to complete the decomposition nevertheless. In this case, we force the decomposition to match the observation data only at the available entries. This is done by introducing a compensation tensor $\mathcal{O}\in \mathbb{R}^{I_1 \times I_2 \times\dots \times I_N}$, which is zero for entries in the observation set $\Omega \subset [I_1] \times [I_2] \times \dots \times [I_N]$, and can take any value outside $\Omega$. Thus using the same auxiliary variables technique as in~\eqref{eq:TPCA_batch}, the problem is formulated as
\begin{equation}
\label{eq:TC_batch}
\begin{aligned}
& \underset{\mathcal{X}_i,\mathcal{E}}{\text{min}}
& & \sum_{i=1}^N\|\mathbf{X}_{i(i)}\|_*+ \lambda \|\mathbf{E}_{(1)}\|_{2,1}\\ 
& \text{s.t.}
& & \mathcal{B = X_\mathnormal{i}+E+O}, i = 1,2,\dots,\ N,\\&
& & \mathcal{O}_\Omega = 0,
\end{aligned}
\end{equation}
where $\mathcal{O}_\Omega$ denotes the entries in $\mathcal{O}$ that are observed. Since $\mathcal{O}$ compensates for whatever the value is in the unobserved entries of $\mathcal{B}$, we only need to keep track of the indices of the unobserved entries, and can simply set the unobserved entries of $\mathcal{B}$ to zero.

Problem~\eqref{eq:TPCA_batch} and~\eqref{eq:TC_batch} are usually solved via  \textit{Alternating direction method of multipliers}~(ADMM) methods or Accelerated Proximal Gradient (APG) methods in an iterative manner~\cite{hu2020robust,goldfarb2014robust}. However, in each iteration, to optimize the term $\|\mathbf{X}_{i(i)}\|_*$\ we need to compute the \textit{singular value decomposition} (SVD) of $\mathbf{X}_{i(i)}$ composed of all samples. This both limits the ability for stream processing of the data, and is computationally expensive. In the next section, we develop an online algorithm to address these limitations.

\section{Methods} \label{sec:Method}
In this section, we first pose the online objective function for online higher-order tensor decomposition problem in the presence of fiber outliers, and then provide an efficient algorithm to solve it in the streaming settings. The online algorithm under partial-observation settings follows the same line as the full observation case.  We provide the formulation and the algorithm for partial observations in the Appendix.

\subsection{Problem formulation}
We now develop the objective function for online setting, starting with the batch-based objective function~\eqref{eq:TPCA_batch}. We note that in streaming data settings, for tensors $\mathcal{E,B,X} \in \mathbb{R}^{I_1 \times I_2 \times\dots \times I_N}$, the size of the last dimension $I_N$ grows with time, but we drop the explicit dependency of time in the notation. As we have seen, the major challenge in a batch-based method~\eqref{eq:TPCA_batch} is that the nuclear norm term $\|\mathbf{X}_{i(i)}\|_*$ keeps all samples tightly coupled. In this section we show how we substitute the nuclear norm term with an equivalent form, based on which we can derive an empirical cost function that separates out the loss incurred by each sample, thus allowing computation sequentially in time.

First, in order to facilitate data online processing, we relax the decomposition constraint in~\eqref{eq:TPCA_batch} into a Frobenious norm penalty in the objective function. Thus,~\eqref{eq:TPCA_batch} becomes

\begin{equation}
\label{eq:TPCA_relax}
\begin{aligned}
& \underset{\mathcal{X}_1, \dots, \mathcal{X}_N,\mathcal{E}}{\text{min}}
& & \frac{1}{2}\sum_{i=1}^N \|\mathcal{X_\mathnormal{i}+E-B}\|_F^2 + \lambda_1 \sum_{i=1}^N\|\mathbf{X}_\mathnormal{i(i)}\|_*  + \lambda_2 \|\mathbf{E}_{(1)}\|_{2,1},
\end{aligned}
\end{equation}
where $\lambda_1$ and $\lambda_2$ are weight parameters balancing the costs of low rank and sparsity respectively.


Next, we deal with the the nuclear norm term $\|\mathbf{X}_{i(i)}\|_*$ that couples all samples and prohibits processing the data sequentially. Namely, for each mode $i = 1,2,\dots,\ N$, we substitute $\|\mathbf{X}_{i(i)}\|_*$ with an equivalent form:
\begin{equation} \label{eq: nuclear}
\begin{aligned}
    & \|\mathbf{X}_{i(i)}\|_* = 
      & &  \underset{\mathbf{L}_i, \mathbf{R}_i}{\text{inf}} \left\{ \frac{1}{2}\|\mathbf{L}_i\|^2_F + \frac{1}{2}\|\mathbf{R}_i\|^2_F  \right\} \\
      &  \text{s.t.}
      && \mathbf{X}_{i(i)} = \mathbf{L}_i\mathbf{R}_i^T.
      \end{aligned}
\end{equation}
where we explicitly factorize $\mathbf{X}_{i(i)}$ into $\mathbf{L}_i\in \mathbb{R}^{I_i \times r} , \mathbf{R}_i\in \mathbb{R}^{ I_{{N\setminus i}} \times r}$, and upper bound the rank of  $\mathbf{X}_{i(i)}$ by $r$, with $r \ll \text{min}(I_i,I_{{N\setminus i}})$. $\mathbf{L}_i$ can be seen as a dictionary, where each column represents a basis vector in the mode-$i$ unfolding, and $\mathbf{R}_i$ are the corresponding coefficients of the basis for the samples. Such nuclear norm substitution~\eqref{eq: nuclear} is well established in works including~\cite{recht2010guaranteed,rennie2005fast}. In this way, Problem~\eqref{eq:TPCA_relax} becomes:
\begin{equation}
\label{eq:TPCA_OL}
\begin{aligned}
& \underset{\mathcal{X}_i,\mathcal{E}}{\text{min}}
& &  \frac{1}{2}\sum_{i=1}^N \|\mathcal{X_\mathnormal{i}+E-B}\|_F^2 + \frac{\lambda_1}{2} \sum_{i=1}^N\left(\|\mathbf{L}_i\|^2_F + \|\mathbf{R}_i\|^2_F \right) + \lambda_2 \|\mathbf{E}_{(1)}\|_{2,1}\\ 
& \text{s.t.}
& & \mathbf{X}_{i(i)} = \mathbf{L}_i\mathbf{R}_i^T,  i = 1,2,\dots,\ N.
\end{aligned}
\end{equation}

Substitution of $\mathbf{X}_{i(i)}$ with  $\mathbf{L}_i$ and $\mathbf{R}_i$ and removing the constraint in~\eqref{eq:TPCA_OL}, we arrive at:
\begin{equation}
\label{eq:TPCA_OL_all}
\begin{aligned}
\underset{\mathbf{L}_i, \mathbf{R}_i, \mathcal{E}}{\text{min}}
&  \frac{1}{2}\sum_{i=1}^N \|\mathbf{L}_i\mathbf{R}_i^T +\mathbf{E}_{(i)} - \mathbf{B}_{(i)}\|_F^2 + \frac{\lambda_1}{2} \sum_{i=1}^N\left(\|\mathbf{L}_i\|^2_F + \|\mathbf{R}_i\|^2_F \right) + \lambda_2 \|\mathbf{E}_{(1)}\|_{2,1}.
\end{aligned}
\end{equation}
For the first term in~\eqref{eq:TPCA_OL_all}, we change the Frobenius norm of a tensor in~\eqref{eq:TPCA_OL} into the Frobenius norm of its mode-$i$ unfolding, which does not change the value of the norm. Problem~\eqref{eq:TPCA_OL_all} is not jointly convex in terms of $\mathbf{L}_i$ and $\mathbf{R}_i$, but a locally minimizing solution for~\eqref{eq:TPCA_OL_all} provides a good global solution for the original problem~\eqref{eq:TPCA_batch}, as theoretically proven in~\cite{feng2013online} in the matrix case, and empirically shown in tensor cases in Section~\ref{sec:experiment}.

In online settings, we divide the overall observation tensor into a series of minibatches along the last dimension,  $\mathcal{B} =  [\mathcal{B}^1, \mathcal{B}^2, \dots, \mathcal{B}^M]$, $\mathcal{B}^t \in \mathbb{R}^{I_1 \times I_2 \times\dots \times I_{N-1}\times I_N'}$,where $I_N' = \lfloor{I_N/M} \rfloor$ is the size of last dimension for each sample. In sensor networks, this amounts to processing the data every day when $I_N'= 1$, or every few days when $I_N'> 1$.

Given the series of samples $\mathcal{B}^t$, solving problem~\eqref{eq:TPCA_OL_all} amounts to minimizing the following empirical objective function:
\begin{equation}
\label{eq:TPCA_L}
\begin{aligned}
f_M(\mathbf{L}_i) \triangleq \frac{1}{M}\sum_{t=1}^M\sum_{i=1}^N\left[l(\mathcal{B}^t, \mathbf{L}_i) + \frac{\lambda_1}{2M}\|\mathbf{L}_i \|_F^2\right],
\end{aligned}
\end{equation}
where the loss function for each mini-batch is defined as
\begin{equation}
\label{eq:TPCA_each}
\begin{aligned}
l(\mathcal{B}^t, \mathbf{L}_i)  =  \underset{\mathbf{R}_i^t, \mathcal{E}^t}{\text{min}}
 &\frac{1}{2} \|\mathbf{L}_i\mathbf{R}_i^{t T} +\mathbf{E}_{(i)}^t - \mathbf{B}_{(i)}^t\|_F^2 + \frac{\lambda_1}{2} \|\mathbf{R}_i^t\|^2_F  + \lambda_2 \|\mathbf{E}_{(1)}^t\|_{2,1}.
\end{aligned}
\end{equation}
In minimizing~\eqref{eq:TPCA_L}, we want to find the set of basis $\mathbf{L}_i$ that works across all samples, and that minimizes the accumulated loss of all samples. The loss for each sample under a fixed basis $\mathbf{L}_i$ is calculated via~\eqref{eq:TPCA_each}, where we find the optimal coefficients $\mathbf{R}_i^t$ and the outlier tensor $\mathcal{E}^t$ for each sample to minimize the loss given the basis $\mathbf{L}_i$. 


\subsection{Online tensor RPCA algorithm}

In this section, we develop OLRTR to efficiently solve Problem~\eqref{eq:TPCA_L} online, taking one mini-batch at a time. The OLRTR algorithm is summarized in Algorithm~\ref{alg:OLTPCA}. In an overview, we take an alternative optimization approach to optimize $\mathbf{R}_i^t$, $\mathcal{E}^t$ and $\mathbf{L}_i$. Namely, at the $t$-th time step, after accessing the new sample $\mathcal{B}^t$, we first solve for the corresponding  $\mathbf{R}_i^t$ and $\mathcal{E}^t$, using the $\mathbf{L}_i$ obtained in last step $t-1$. Then we update $\mathbf{L}_i$, to minimize the accumulated loss given all $\{\mathbf{R}_i^\tau\}_{\tau=1}^t$ and $\{\mathcal{E}^\tau\}_{\tau=1}^t$ obtained so far.

Specifically, at time step $t$, having accessed the minibatch $\mathcal{B}^t$, we first address~\eqref{eq:TPCA_each} and solve for $\mathbf{R}_i^t$ and $\mathcal{E}^t$ with $\mathbf{L}_i$ fixed. To this end, we alternatively update $\mathbf{R}_i^t$ and $\mathcal{E}^t$. When $\mathcal{E}^t$ is fixed, we solve $\mathbf{R}_i^t$ via: 
\begin{equation}
\label{eq:R}
\mathbf{R}_i^t= \underset{\mathbf{R}_i}{\text{argmin }}
\frac{\lambda_1}{2} \|\mathbf{R}_i\|^2_F + \frac{1}{2} \sum_{i=1}^N \|\mathbf{L}_i\mathbf{R}_i^{T} + \mathbf{E}_{(i)}^t - \mathbf{B}_{(i)}^t\|_F^2,
\end{equation}
which has a closed-form solution
\begin{equation}
\label{eq:R2}
\mathbf{R}_i^t = \left(\mathbf{L}_i^T \mathbf{L} + \lambda_1\mathbf{I} \right)^{-1}\mathbf{L}^T\left( \mathbf{B}_{(i)} - \mathbf{E}_{(i)} \right).
\end{equation}
Then, when $\mathbf{R}_i^t$ is fixed, we solve $\mathcal{E}^t$ via  
\begin{equation}
\begin{aligned}
\label{eq:E}
\mathcal{E}^t &= \underset{\mathcal{E}}{\text{argmin }}
\lambda_2 \|\mathbf{E}^t_{(1)}\|_{2,1} + \frac{1}{2} \sum_{i=1}^N \|\mathbf{E}_{(i)} + \mathbf{L}_i\mathbf{R}_i^{t T}  - \mathbf{B}_{(i)}^t\|_F^2 \\
 &= \underset{\mathcal{E}}{\text{argmin }}
\lambda_2  \|\mathbf{E}_{(1)}\|_{2,1}  + \frac{1}{2} \sum_{i=1}^N \|\mathcal{E} + \text{fold}_i(\mathbf{L}_i\mathbf{R}_i^{t T})   - \mathcal{B}^t\|_F^2 
\end{aligned}
\end{equation}
Following the same approach as~\cite{goldfarb2014robust, hu2020robust}, problem~\eqref{eq:E} shares the same solution as
\begin{equation}
\begin{aligned}
\label{eq:E2}
\mathcal{E}^t= \underset{\mathcal{E}}{\text{argmin }}
& \lambda_2 \|\mathbf{E}_{(1)}\|_{2,1} +  \frac{N}{2} \left\rVert\mathcal{E} - \frac{1}{N} \sum_{i=1}^N\left(\mathcal{B}^t -\text{fold}_i(\mathbf{L}_i\mathbf{R}_i^{t T}) \right) \right\rVert_F^2.    
\end{aligned}
\end{equation}
Denoting the term $\left( \mathcal{B}^t -\text{fold}_i(\mathbf{L}_i\mathbf{R}_i^{t T})\right)$ as $\mathcal{C}$, then following the approach in~\cite{hu2020robust}, the closed form solution for~\eqref{eq:E} is:
\begin{equation}
\begin{aligned}
\mathbf{E}_{(1)j}^{t} = \mathbf{C}_{(1)j}~\text{max}\left\{0,1-\frac{\lambda_2}{\mu N\|\mathbf{C}_{(1)j}\|_2}\right\}, \text{ for } j = 1,2,\dots ,p,
 \end{aligned}
\end{equation}
where $\mathbf{E}_{(1)j}$ is the $j^{th}$ column of $\mathbf{E}_{(1)}$,  $\mathbf{C}_{(1)j}$ is the $j^{th}$ column of  $\mathbf{C}_{(1)}$, and $p = I_2 \times I_3 \times \dots \times I_N'$ is the total number of columns in $\mathbf{C}_{(1)}$.

The sample update for $\mathbf{R}_i$ and $\mathcal{E}$ is summarized in Algorithm~\ref{alg:R_E}. The algorithm 
convergence criterion is met when the change between iterations is small enough, as measured by the Frobenious norm of the difference in $\mathbf{R}_i$ and $\mathcal{E}$ between iterations i.e.,
\begin{equation}
\label{eq:converge}
     \text{max}\left(\frac{\|\mathbf{R}_i^{(k-1)} - \mathbf{R}_i^{(k)}\|_F}{\|\mathcal{B}^t \|_F} , \frac{\|\mathcal{E}^{(k-1)} - \mathcal{E}^{(k)}\|_F}{\|\mathcal{B}^t \|_F} \right) \leq \epsilon,
\end{equation}
where $(k)$ denotes the iteration number, and $\epsilon$ is the tolerance. 

Next, given the estimated coefficient $\mathbf{R}_i^t$ and outlier tensor $\mathcal{E}^t$, we update the dictionary $\mathbf{L}_i^t$. We define the objective function for updating $\mathbf{L}_i^t$ as 
\begin{equation}
\label{eq:L}
\begin{aligned}
g_t(\mathbf{L}_i)  \triangleq  & \frac{1}{t}\sum_{t=1}^t \big( \left\rVert\mathbf{L}_i\mathbf{R}_i^{t T} +\mathbf{E}_{(i)}^t - \mathbf{B}_{(i)}^t\right\rVert_F^2  + \frac{\lambda_1}{2} \|\mathbf{R}_i^t\|^2_F  + \lambda_2 \|\mathbf{E}_{(1)}^t\|_{2,1} \big) + \frac{\lambda_1}{2t} \|\mathbf{L}_i\|^2_F,
\end{aligned}
\end{equation}
and we aim to solve
\begin{equation}
    \label{eq:L2}
    \mathbf{L}_i^t = \underset{\mathbf{L}_i}{\text{argmin }} g_t(\mathbf{L}_i).
\end{equation}
$g_t(\mathbf{L}_i)$ is a surrogate function for $f_N(\mathbf{L}_i)$ in~\eqref{eq:TPCA_L}, and provides an upper bound for $f_N(\mathbf{L}_i)$. Using the relationship between the Frobenius norm and the matrix trace, $\|\mathbf{Y}\|^2_F = \text{Tr}(\mathbf{Y}^T\mathbf{Y})$, and the properties of matrix trace computation, Problem~\eqref{eq:L2} can be transformed into
\begin{equation}
\label{eq:L3}
\begin{aligned}
\mathbf{L}_i^t = \underset{\mathbf{L}_i}{\text{argmin }} \frac{1}{2}\text{Tr}\left(\mathbf{L}_i^T \left(\mathbf{A}_i^t+\lambda_1\mathbf{I}\right)\mathbf{L}_i  \right) - \text{Tr}\left(\mathbf{L}_i^T\mathbf{D}_i^t \right),
\end{aligned}
\end{equation}
where the two sets of accumulation matrices $\mathbf{A}_i^t \in \mathbb{R}^{r \times r}$ and $\mathbf{D}_i^t \in \mathbb{R}^{I_i \times r}$ are defined as 
\begin{equation*}
\begin{aligned}
    &\mathbf{A}_i^t = \sum_{t =1}^t \mathbf{R}_i^{t T}\mathbf{R}^t_i, \\
   & \mathbf{D}_i^t = \sum_{t=1}^t \left(\mathbf{B}^t_{(i)} - \mathbf{E}^t_{(i)} \right)\mathbf{R}^t_i.
\end{aligned}
\end{equation*}

To solve Problem~\eqref{eq:L3}, we adopt a block-coordinate decent approach similar to~\cite{feng2013online,mairal2010online}, updating one column of $\mathbf{L}_i$ at a time with the rest columns fixed. At each step $t$, we use the solution for the previous step, $\mathbf{L}_i^{t-1}$ as warm restart. The algorithm is provided in Algorithm~\ref{alg:Li}. In online computation, we store the value of $\mathbf{A}_i^t $ and $\mathbf{D}_i^t $ to accumulate the information in all samples, whose sizes does not change with the number of samples, thus enabling the scalabilty of the online algorithm.

We initialize each entry of the dictionary $\mathbf{L}_i \in \mathbb{R}^{I_i \times r}$ from an i.i.d. uniform distribution $\mathcal{U}(0,1)$, and initialize the accumulation matrices $\mathbf{A}_i^t \in \mathbb{R}^{r \times r}, \mathbf{D}_i^t \in \mathbb{R}^{I_i \times r}$ to zero.

\begin{algorithm}[t]
  \caption{OLRTR algorithm}\label{alg:OLTPCA}
  \begin{algorithmic}[1]
     \State Given $N$-way minibatch tensors $[\mathcal{B}^1, \mathcal{B}^2, \dots, \mathcal{B}^M]$ with $\mathcal{B}^t \in \mathbb{R}^{I_1 \times I_2 \times\dots \times I_{N-1}\times I_N'}$, weighting parameters $\lambda_1, \lambda_2$, target rank $r$. Initialize dictionary $\mathbf{L}_i \in \mathbb{R}^{I_i \times r}$ and accumulation matrices $\mathbf{A}_i^t \in \mathbb{R}^{r \times r}, \mathbf{D}_i^t \in \mathbb{R}^{I_i \times r}$.
     \For{$t = 0,1,\dots , M$} 
        \State Access the $t$-th sample $\mathcal{B}^t$
        \State Solve $\mathbf{R}_i^t$ and $\mathcal{E}^t$ for the new sample using Algorithm~\ref{alg:R_E}.
        \begin{equation*}
        \begin{aligned}
            \{ \mathbf{R}_i^t, \mathcal{E}^t \} = \underset{\mathbf{R}_i^t, \mathcal{E}^t}{\text{argmin}}
 \frac{1}{2}\sum_{i=1}^N \|\mathbf{L}_i\mathbf{R}_i^{t T} +\mathbf{E}_{(i)}^t - \mathbf{B}_{(i)}^t\|_F^2 + \\ \frac{\lambda_1}{2} \sum_{i=1}^N \|\mathbf{R}_i^t\|^2_F  + \lambda_2 \|\mathbf{E}_{(1)}^t\|_{2,1}.
        \end{aligned}
        \end{equation*} 
        \State $\mathcal{X}^t = \frac{1}{N} \text{fold}_i \left(\mathbf{L}_i\mathbf{R}_i^t \right)$
        \State $A_i^t = A_i^{t-1} + \mathbf{R}_i^{t T}\mathbf{R}^t_i$;  $ \mathbf{D}_i^t =\mathbf{D}_i^{t-1} + \left(\mathbf{B}^t_{(i)} - \mathbf{E}^t_{(i)} \right)\mathbf{R}^t_i$
        \State Solve $\mathbf{L}_i^t$ with Algorithm~\ref{alg:Li}, warm started with $\mathbf{L}_i^{t-1}$
        \begin{equation*}
            \mathbf{L}_i^t = \underset{\mathbf{L}_i}{\text{argmin }} \frac{1}{2}\text{Tr}\left(\mathbf{L}_i^T \left(\mathbf{A}_i^t+\lambda_1\mathbf{I}\right)\mathbf{L}_i  \right) - \text{Tr}\left(\mathbf{L}_i^T\mathbf{D}_i^t \right)
        \end{equation*}
    \EndFor
    \State \Return low rank tensors $[\mathcal{X}^1, \mathcal{X}^2, \dots, \mathcal{X}^M]$ and outlier tensors $[\mathcal{E}^1, \mathcal{E}^2, \dots, \mathcal{E}^M]$ 
  \end{algorithmic}
\end{algorithm}

\begin{algorithm}[t]
  \caption{Sample update for $\mathbf{R}_i$ and $\mathcal{E}$}\label{alg:R_E}
  \begin{algorithmic}[1]
     \State Given observation $\mathcal{B}  \in \mathbb{R}^{I_1 \times I_2 \times\dots \times I_{N-1}\times I_N'}$, basis $\mathbf{L}_i  \in \mathbb{R}^{I_i \times r}$ and parameters $\lambda_1, \lambda_2$. Initialize coefficient matrix $\mathbf{R}_i \in \mathbb{R}^{I_{N \setminus i}' \times r}$ and outlier tensor $\mathcal{E} \in \mathbb{R}^{I_1 \times I_2 \times\dots \times I_{N-1}\times I_N'}$ to zero
    \While {not converged}
        \State $\mathcal{C} \gets \left( \mathcal{B} -\text{fold}_i(\mathbf{L}_i\mathbf{R}_i^{ T})\right)$ \Comment{Update $\mathcal{E}.$ }
        \For{$j = 1,2 ,\dots, p$}
        \State  $\mathbf{E}_{(1)j} \gets \mathbf{C}_{(1)j}\text{max}\left\{0,1-\frac{\lambda}{\mu N\|\mathbf{C}_{(1)j}\|_2}\right\}$
        \EndFor
        \For{$i = 1,2 ,\dots, N$} \Comment{Update $\mathbf{R}_i$}
        \State $\mathbf{R}_i \gets \left(\mathbf{L}_i^T \mathbf{L} + \lambda_1\mathbf{I} \right)^{-1}\mathbf{L}^T\left( \mathbf{B}_{(i)} - \mathbf{E}_{(i)} \right).$ 
        \EndFor
    \EndWhile
    \State \Return $\mathbf{R}_i$ and $\mathcal{E}$ 
  \end{algorithmic}
\end{algorithm}

\begin{algorithm}[t]
  \caption{Basis $\mathbf{L}_i$ update}\label{alg:Li}
  \begin{algorithmic}[1]
     \State Given $\mathbf{L}_i = [\mathbf{l}_{i\_1}, \dots, \mathbf{l}_{i\_r}] \in \mathbb{R}^{I_{N\setminus i}\times r}, \mathbf{A}_i =  [\mathbf{a}_{i\_1}, \dots, \mathbf{a}_{i\_r}]  \in \mathbb{R}^{r \times r}, \mathbf{D}_i =  [\mathbf{d}_{i\_1}, \dots, \mathbf{d}_{i\_r}]  \in \mathbb{R}^{I_i \times r} $. Let $\Tilde{\mathbf{A}}_i = \mathbf{A}_i + \lambda_1\mathbf{I}$.
     \For{$j = 0,1,\dots r$}
        \State $\mathbf{l}_{i \_ j} \gets \frac{1}{\Tilde{\mathbf{A}}_{i\_j,j}} \left( \mathbf{d}_{i\_j} - \mathbf{L}_i\mathbf{a}_{i\_j}  \right) + \mathbf{l}_{i \_ j}$
    \EndFor
    \State \Return $ \mathbf{L}_i$
  \end{algorithmic}
\end{algorithm}

\subsection{Complexity and memory cost}

The overall complexity for Algorithm~\ref{alg:OLTPCA} depends on the minibatch size. Namely, denoting the overall size for a minibatch $\mathcal{B}^t \in \mathbb{R}^{I_1 \times I_2 \times\dots \times I_{N-1}\times I_N'}$ as $s = \prod_{i=1}^{N}I_i =  I_1 \times I_2 \times \dots \times I_{N-1}\times I_N'$, then the computational complexity is $O(rs)$. To see this, we note that the complexity for line 4 in Algorithm~\ref{alg:OLTPCA} is $O(r^2I_i+r^3+rs)$ using Algorithm~\ref{alg:R_E}, where the $O(r^3)$ comes from the matrix inverse, and $O(r^2I_i+rs)$) comes from matrix multiplications. The complexity for line 6 is $O(r^2I_{N\setminus i} + rs)$. The complexity for line 7 is $O(r^2I_i)$ with Algorithm~\ref{alg:Li}, since to update each column of $L_i$ takes $O(rI_i)$, and there are $r$ columns in total. Since $r \ll I_i < s$, the overall complexity is thus $O(rs)$, which is linear with the minibatch size, and is relatively small with a reasonable minibatch size.

The memory cost for Algorithm~\ref{alg:OLTPCA} in each iteration is $O(s)$, dominated by loading the minibatch data $\mathcal{B}^t$ and estimating $\mathcal{X}^t$ and $\mathcal{E}^t$. The historical information has been stored in $\mathbf{A}^t$ and $\mathbf{D}^t$, at a memory cost of $O(rI_i)$. Thus, the memory cost does not increase with the number of samples, meeting the need for large-scale long term monitoring systems.


\section{Experiments}
\label{sec:experiment}

In this section, we conduct experiments on both simulation data and real world sensor network data. We first examine the performance of tensor recovery and anomaly detection on synthetically generated low-rank tensors with known fiber-wise corruptions and random missing entries. Then we apply the algorithm on two large sensor network datasets. The first experiment is a complete NOAA~\cite{noaa_data} temperature dataset that we synthetically degrade, so that the recovery relative error can be computed. In the second sensor network experiment, we apply the method to Array of Things temperature data which contains missing data and outliers. We assess the quality of the recovery by comparing the correlation of AoT data with NOAA sensors when they are in close proximity.   

We compare our method with online tensor approaches including ORLTM~\cite{li2019online}, OSTD~\cite{sobral2015online}, online matrix approaches OLRSC~\cite{shen2016online}, STOC-RPCA~\cite{feng2013online}, GRASTA~\cite{he2011online}, as well as batch tensor approaches TRPCA~\cite{lu2019tensor}, and RTR~\cite{hu2020robust}.

The performance of the algorithms are measured by the relative error of the low rank tensor, as well as the F1 score of the outlier fibers. The \textit{relative error} (RE) of low rank tensor is calculated as:
\begin{equation}
    \label{eq:re}
   \text{RE} = \frac{\|\mathcal{X}_{0\text{all}}-\hat{\mathcal{X}}_\text{all}\|_F}{\|\mathcal{X}_{0\text{all}}\|_F},
\end{equation}
where $\| \cdot \|_F$ is the tensor Frobenius norm, and $\hat{\mathcal{X}}_\text{all} \in \mathbb{R}^{I_1 \times I_2 \times nI_3}$ is the estimated low rank tensor, which has the value 0 in the fibers that are estimated to be corrupted.  For online methods, $\hat{\mathcal{X}}_\text{all}$ is constructed by concatenating the estimated low rank tensors for all samples along the last dimension. Since most online algorithms are cold started, leading to large losses when first initialized, we discard the first 10 minibatch samples and only compare the performance on the remaining samples. 

For our proposed model OLRTR, we set $\epsilon = 10^{-4}$, and we use an empirical value $\lambda_1 = 0.01$, $\lambda_2 = \frac{\alpha}{\sqrt{\text{log}(I_mI_m)}}$
where $I_m = \text{max}(I_1, \dots , I_N)$, and $\alpha$ is a parameter to tune. For other methods, we use the default values as indicated in the original papers, or tuned for best results if the default values doesn't work. The code and data for experiments can be found in \url{https://github.com/yuehu9/Online_Robust_Tensor_Recovery}.

 \subsection{Numerical experiments}
 
In this section, we conduct a series of numerical experiments to examine the performance of our method on synthetically generated datasets. 
 
\subsubsection{Simulation setup}
We synthetically generate a series of minibatch observation data ($\mathcal{B}^1, \mathcal{B}^2, \cdots, \mathcal{B}^M$), where $\mathcal{B}^t$ is the $t$-th minibatch sample, and is generated as $\mathcal{B}^t = \mathcal{X}_0^t + \mathcal{E}_0 ^t\in  \mathbb{R}^{I_1 \times I_2 \times I_3}$. The ground truth data $\mathcal{X}_0^t$ is generated with a core tensor $\mathcal{G}^t \in \mathbb{R}^{c_1 \times c_2 \times c_3}$ multiplied in each mode by orthogonal matrices of corresponding dimensions, $\mathbf{U}^{(i)} \in \mathbb{R}^{I_i \times c_i}$, i.e.,  $
      \mathcal{X}_0^t= \mathcal{G}^t \times_1 \mathbf{U}^{(1)}\times_2 \mathbf{U}^{(2)} \times_3 \mathbf{U}^{(3)}.
$
The entries of $\mathcal{G}^t$ are independently sampled from standard Gaussian distribution. The orthogonal matrices $\mathbf{U}^{(i)}$ are generated via a Gram-Schmidt orthogonalization on $c_i$ vectors of size $\mathbb{R}^{I_i}$ drawn from standard Gaussian distribution. $\mathbf{U}^{(i)}$ are kept the same across $M$ minibatch samples, i.e., do not change with $t$, so that all minibatches share the same low rank basis. The fiber sparse tensor $\mathcal{E}_0^t$ is formed by first generating a tensor $\mathcal{E}_0'^t \in \mathbb{R}^{I_1 \times I_2 \times I_3}$, whose entries are i.i.d uniform distribution $\mathcal{U}$(-2,2). Then  we randomly keep a fraction $\gamma$ of the fibers of $\mathcal{E}_0'^t$ to form $\mathcal{E}_0^t$. Finally, the corresponding fibers of $\mathcal{X}_0^t$ with respect to non-zero fibers in $\mathcal{E}_0^t$ are set to zero.
For batch methods, we concatenate all samples $\mathcal{B}^1, \mathcal{B}^2, \cdots, \mathcal{B}^M$ along the last dimension to form the observation  $\mathcal{B}_\text{all} \in \mathbb{R}^{I_1 \times I_2 \times MI_3}$ with low rank and sparse components $\mathcal{X}_{0\text{all}}, \mathcal{E}_{0\text{all}} \in \mathbb{R}^{I_1 \times I_2 \times MI_3}$. For matrix-based online methods, we unfold $\mathcal{B}_\text{all}$ in first mode and feed the matrix into the algorithms. In regards of hyper-parameters, we set $\alpha=3$ for OLRTR, and set the rank upper bound $c$ as the true rank for all methods. 

\subsubsection{Experiment results}
\begin{figure}
\centering
\begin{subfigure}[t]{0.45\columnwidth}
    \includegraphics[width=\linewidth]{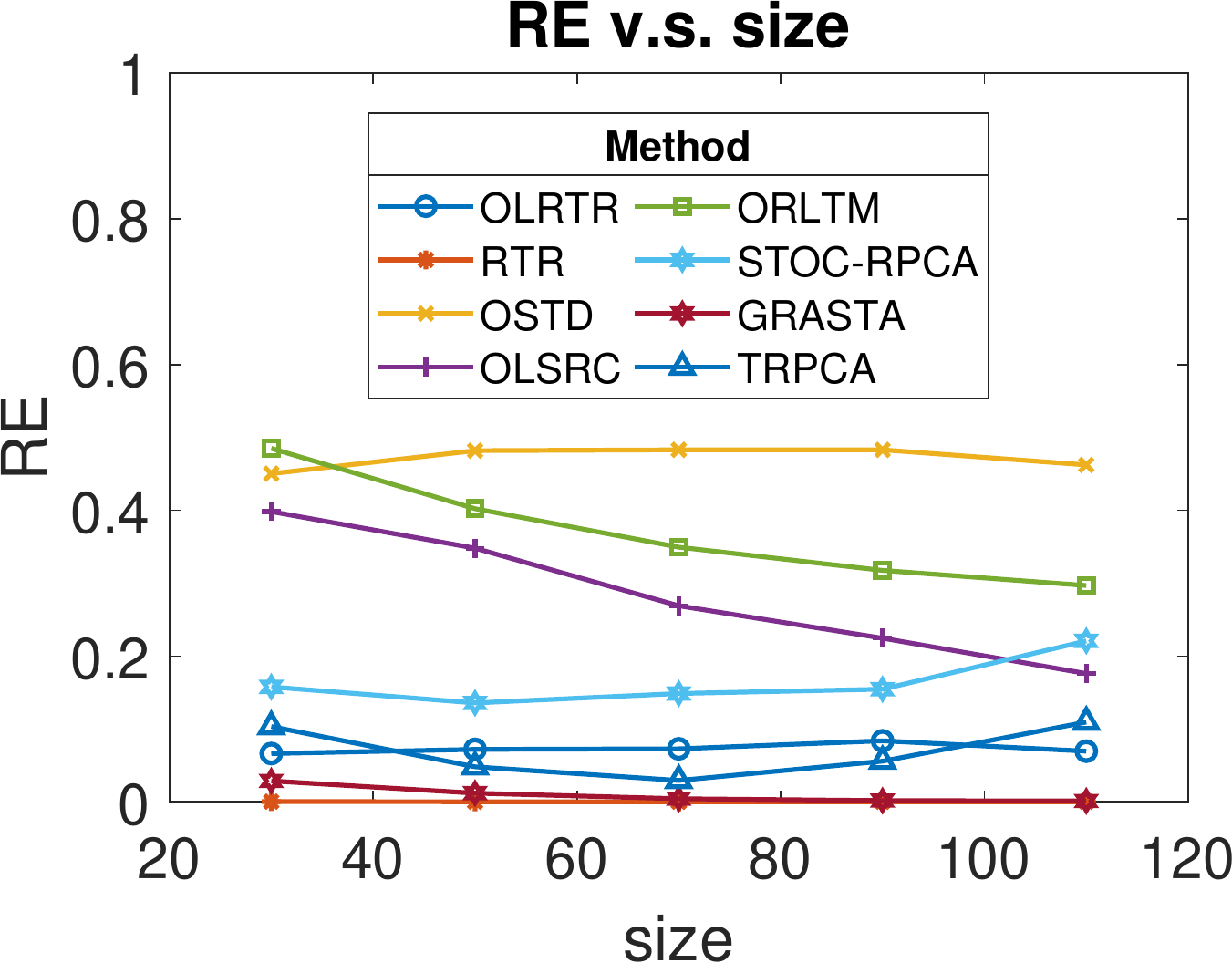}
    \caption{}
    \label{fig:traiffic_graph1}
\end{subfigure}
\quad
\begin{subfigure}[t]{0.45\columnwidth}
    \includegraphics[width=\linewidth]{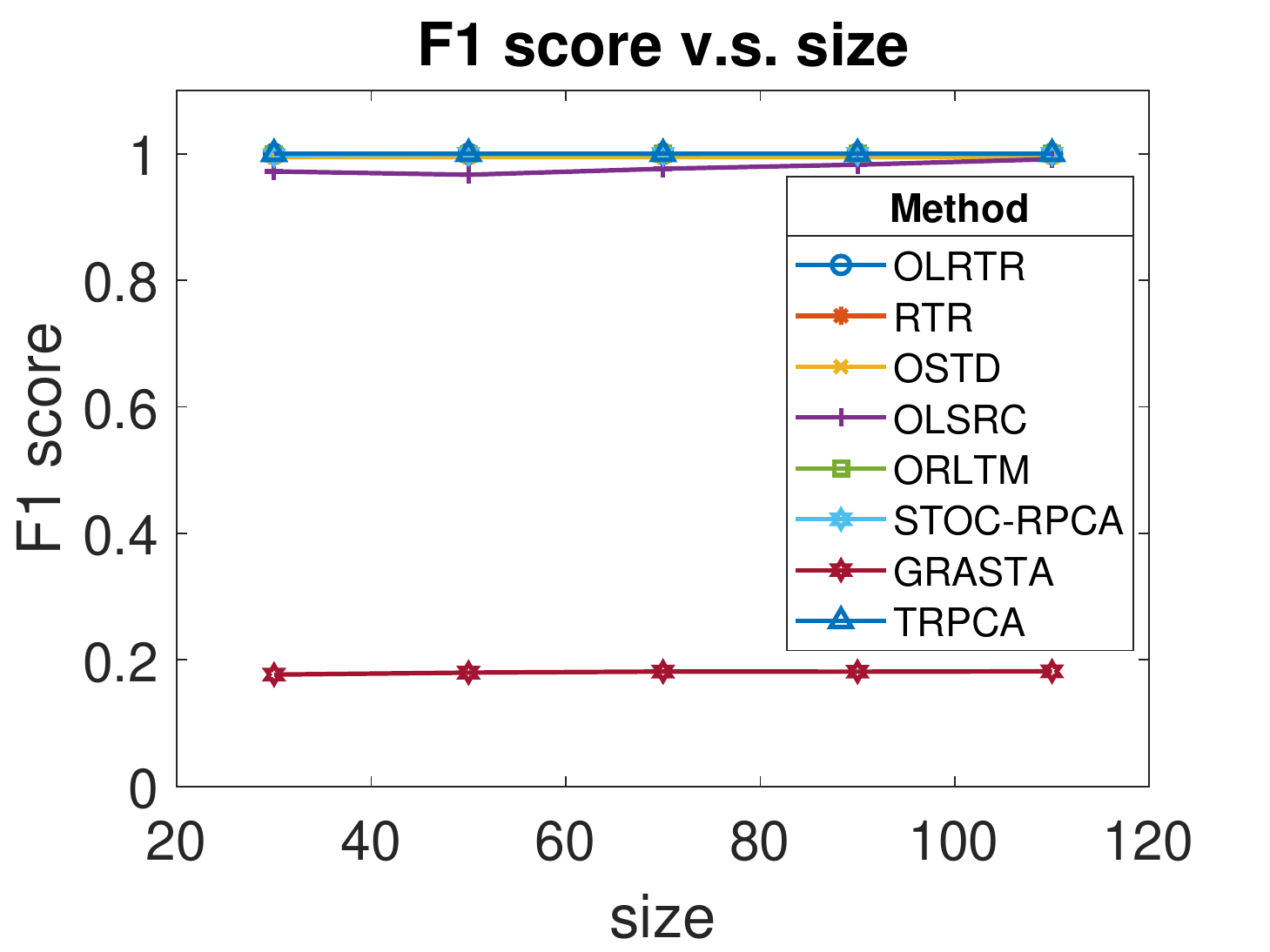}
    \caption{}
    \label{fig:adj1}
\end{subfigure}
\quad
\begin{subfigure}[t]{0.45\columnwidth}
    \includegraphics[width=\linewidth]{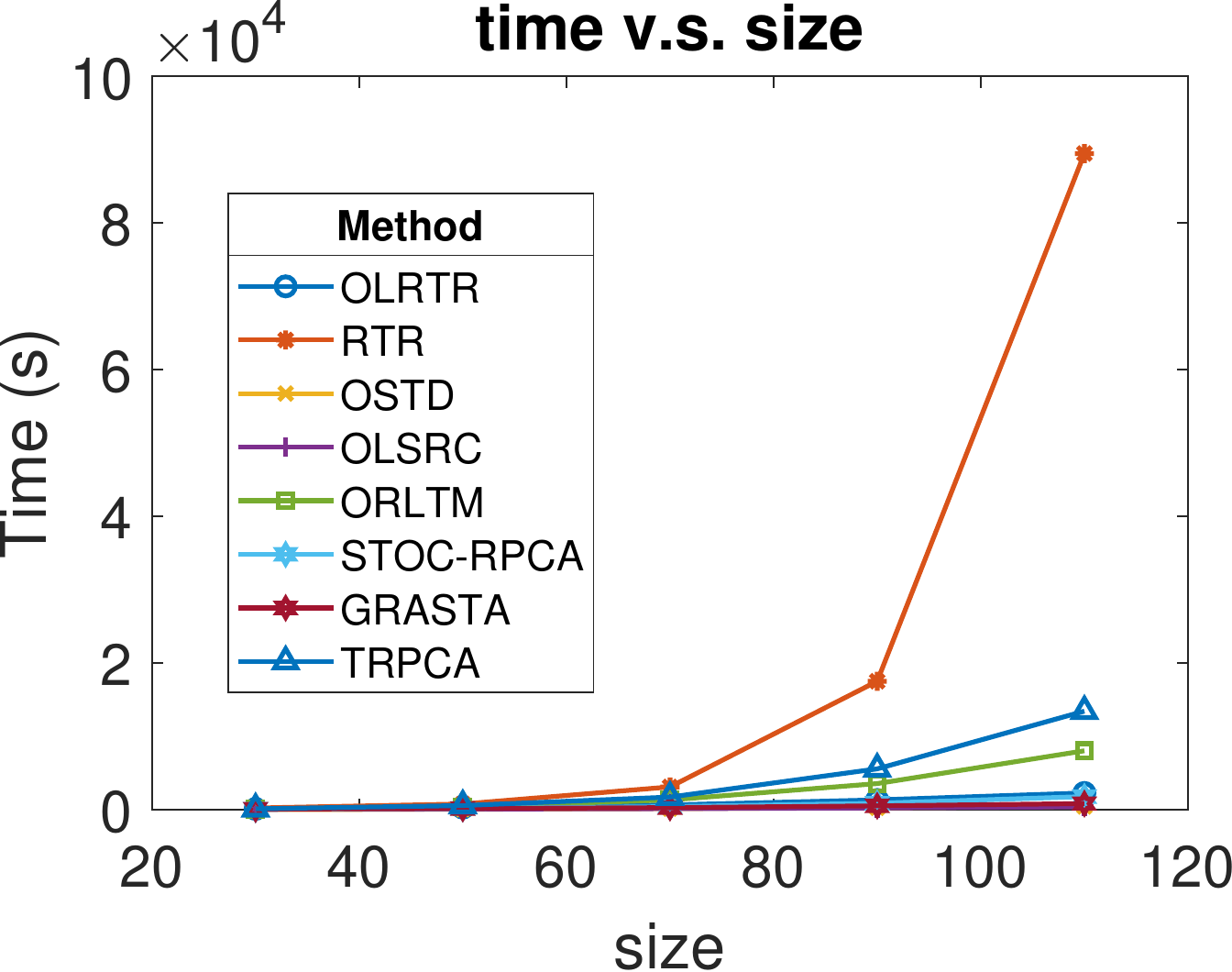}
    \caption{}
    \label{fig:adj2}
\end{subfigure}
\quad
\begin{subfigure}[t]{0.45\columnwidth}
    \includegraphics[width=\linewidth]{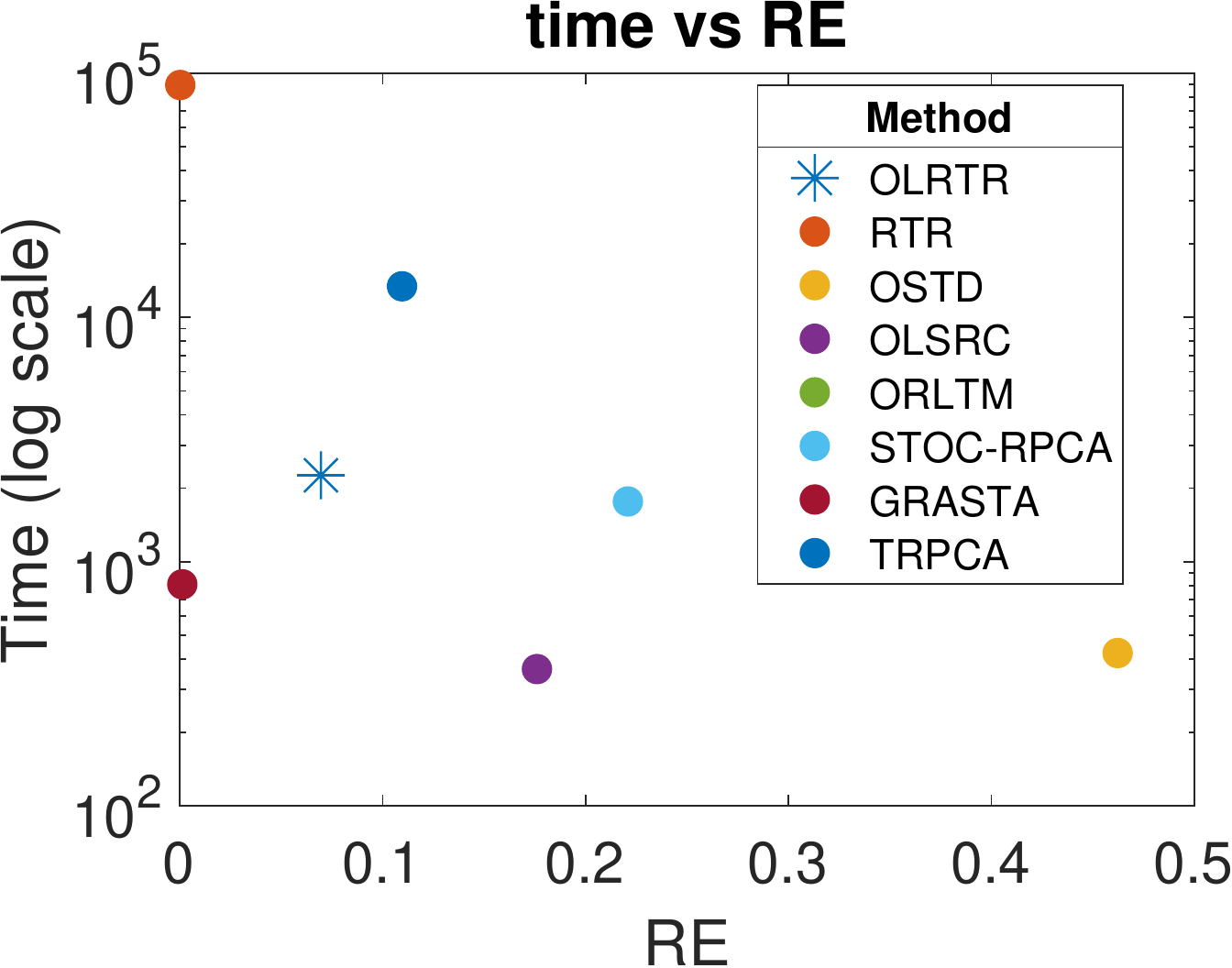}
    \caption{}
    \label{fig:adj3}
\end{subfigure}
\caption{Results of algorithms as a function of tensor size $I$. For each experiment, we run 100 minibatches, varying the tensor size of each minibatch $\mathbb{R}^{I \times I \times I}$. The rank is set at $0.1I$, the gross corruption ratio $\gamma$ is set at 0.1, and the observation ratio is set at 0.9. The result is an average over 5 trials.}
\label{fig:size}
\end{figure}

First, we vary the tensor size, and compare the performance in terms of residual error, F1 score and time. For each experiment, we run 100 minibatches, varying the tensor size of each minibatch $\mathbb{R}^{I \times I \times I}$. The rank is set at $0.1I$ and the gross corruption ratio $\gamma$ is set at 0.1, and the observation ratio is set at 0.9.
ime The result is shown in Fig.~\ref{fig:size}. We can see that only batch approach RTR~\cite{hu2020robust} can exactly recover the low rank tensor and find all outliers, yet its computation time increases sharply as input size increases, showing that RTR is not scalable for large online systems. Our algorithm OLRTR works the second best considering RE and F1 score, comparable with the batch method TRPCA~\cite{lu2019tensor}. GRASTA~\cite{he2011online} performs well in terms of computation time and RE, but has low F1 score, meaning it is lacking in detecting outliers. STOC-RPCA~\cite{feng2013online} is the matrix counterpart to our method for online settings with element-wise outliers, and our proposed method has about a 0.1 improvement in RE over the STOC-RPCA method. This shows the advantage of our tensor approach in taking full advantage of the correlations in every dimension, to get the best estimate for the low rank tensor.

\begin{figure}
\centering
\begin{subfigure}[t]{0.45\columnwidth}
    \includegraphics[width=\linewidth]{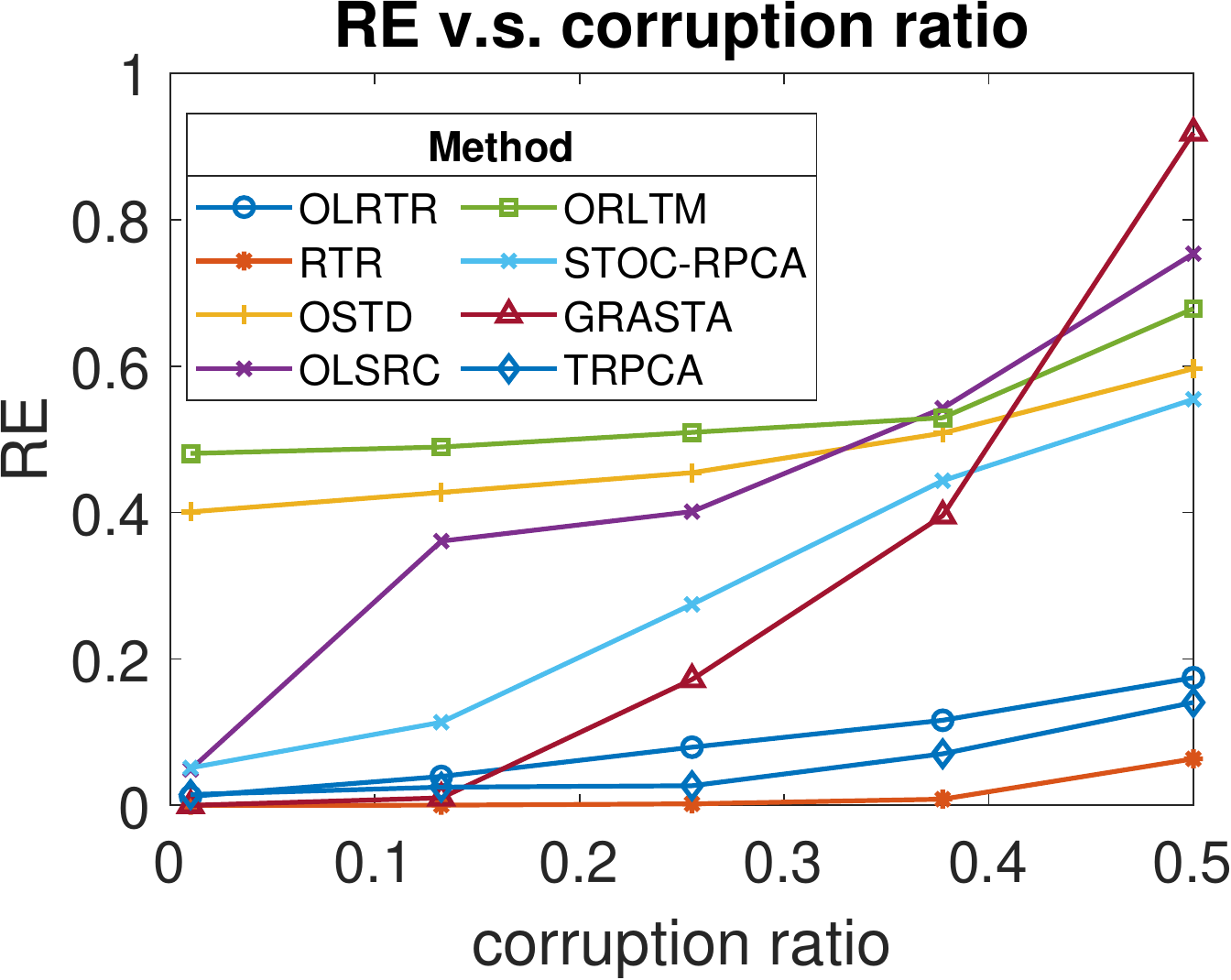}
    \caption{}
    \label{fig:traiffic_graph2}
\end{subfigure}
\quad
\begin{subfigure}[t]{0.45\columnwidth}
    \includegraphics[width=\linewidth]{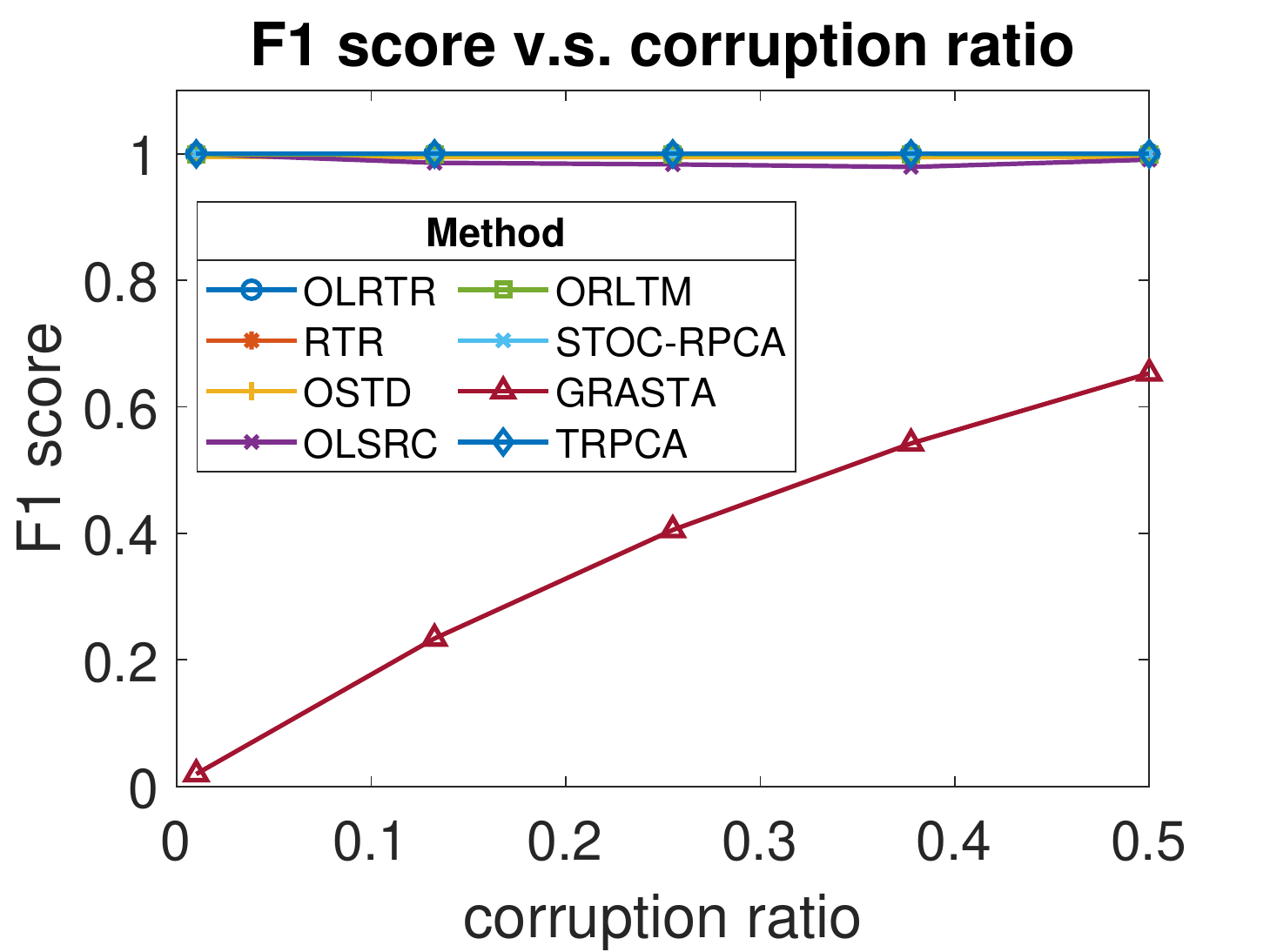}
    \caption{}
    \label{fig:adj4}
\end{subfigure}

\begin{subfigure}[t]{0.45\columnwidth}
    \includegraphics[width=\linewidth]{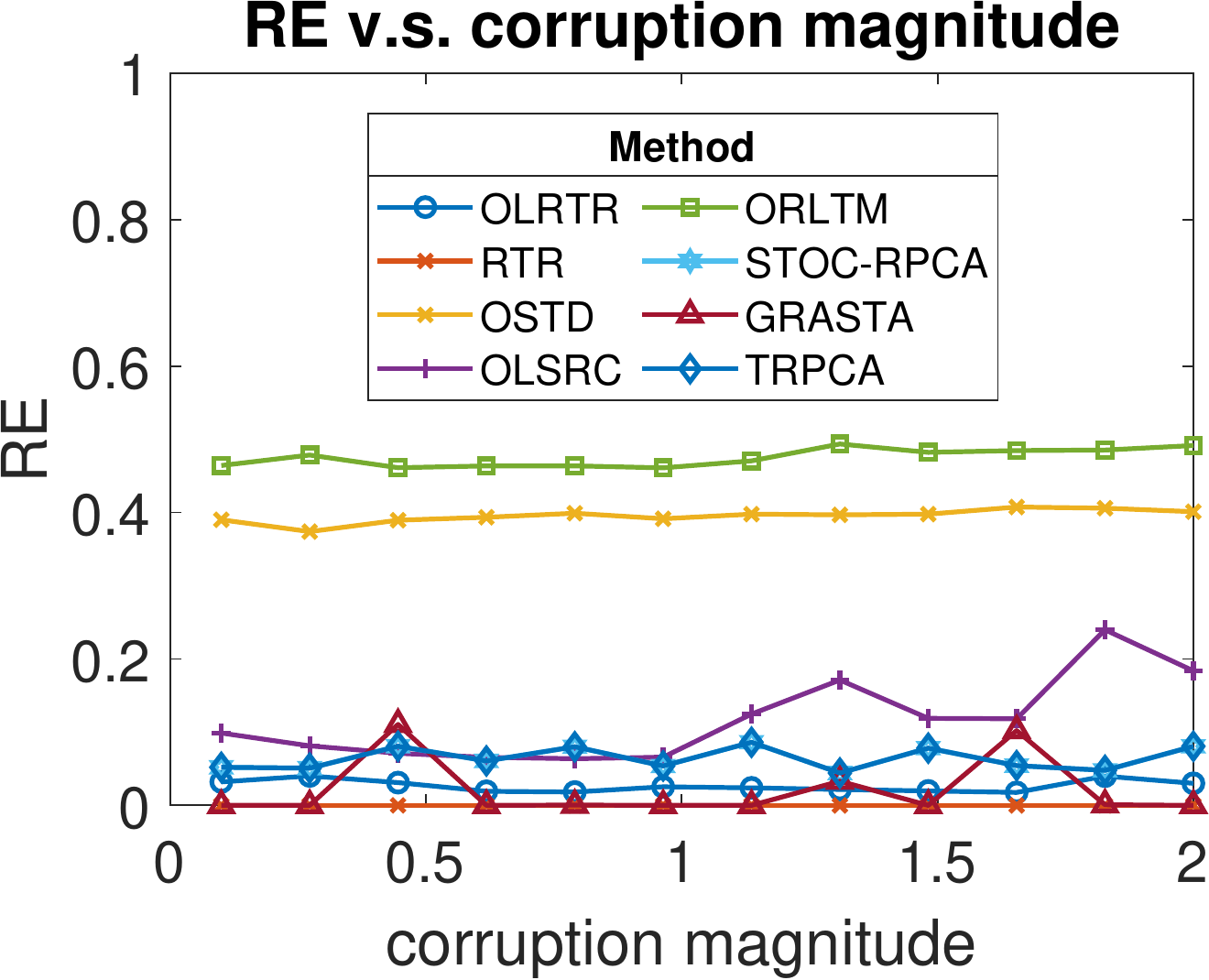}
    \caption{}
    \label{fig:adj5}
\end{subfigure}
\quad
\begin{subfigure}[t]{0.45\columnwidth}
    \includegraphics[width=\linewidth]{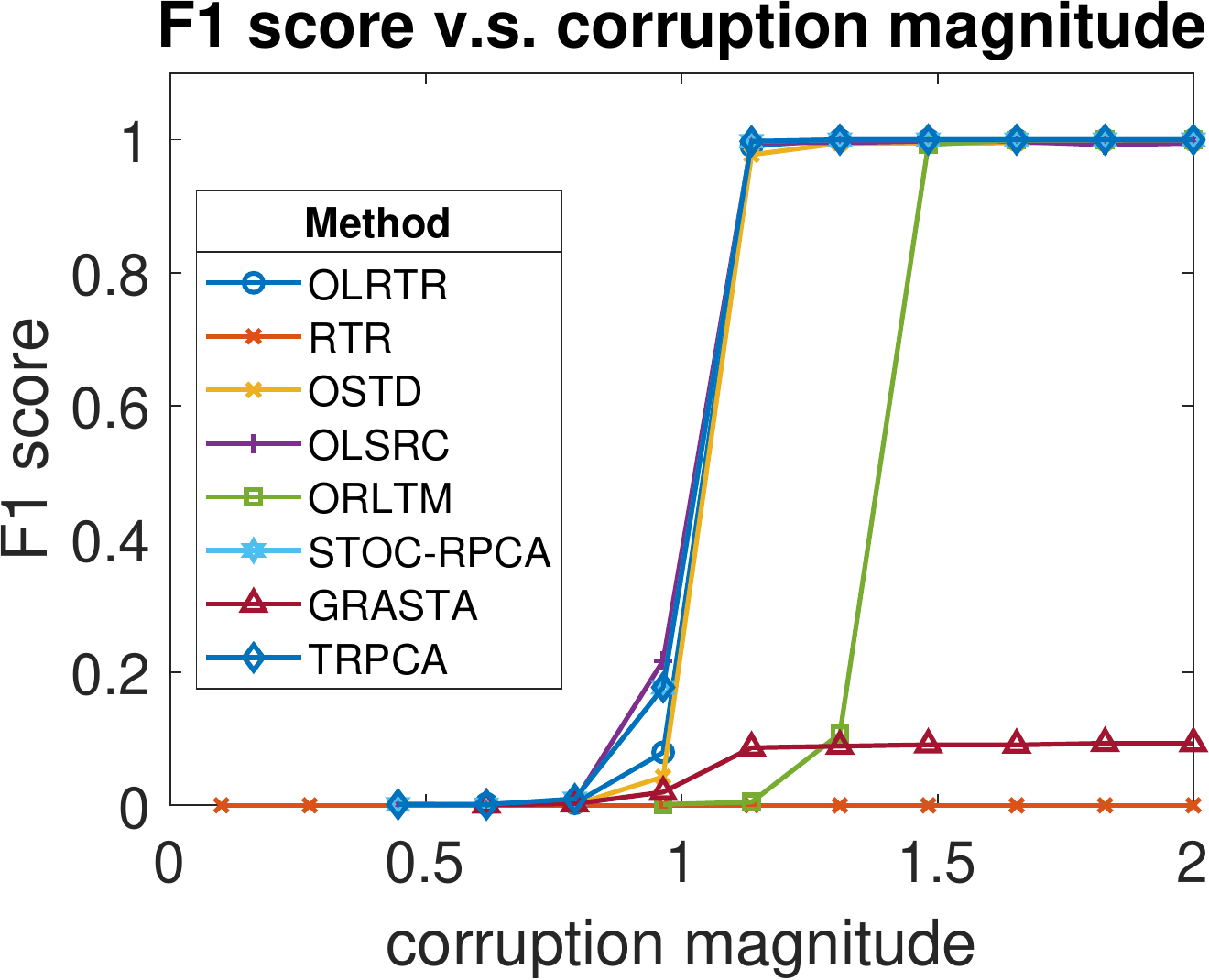}
    \caption{}
    \label{fig:adj6}
\end{subfigure}
\caption{Results of algorithms with varying corruption ratio (upper row) and magnitude(lower row). We run 100 minibatches. The low-rank tensor size for each minibatch is fixed at $\mathbb{R}^{50 \times 50 \times 50}$ with a tucker rank of $(3,3,3)$, and the observation ratio is set at 1. The result is an average over 5 trials.}
\label{fig:noise}
\end{figure}

Next, we vary the fiber corruption ratio and magnitude, and investigate the residual error and F1 score. We run 100 minibatches at each corruption ratio. The low-rank tensor size for each minibatch is fixed at $\mathbb{R}^{50 \times 50 \times 50}$ with a tucker rank of $(3,3,3)$, and the observation ratio is set at 1. The result is shown in Fig.~\ref{fig:noise}. We can see that as corruption ratio increases from 0 to 0.5, RE increases for all methods. But the relative error for OLRTR is always under 0.2, second only to the batch methods RTR and TRPCA. The performance of all other online methods drop sharply, with RE above 0.5 for a corruption ratio $\gamma$ of 0.5. We also vary the corruption magnitude with a fixed corruption ratio $\gamma = 0.05$. From the second row of Fig.~\ref{fig:noise}, we see that the relative error is not sensitive to the corruption magnitude. However, the F1 score shows that no method detects outliers if the corruption magnitude is sufficiently small. 

\begin{figure}
\centering
\begin{subfigure}[t]{0.45\columnwidth}
    \includegraphics[width=\linewidth]{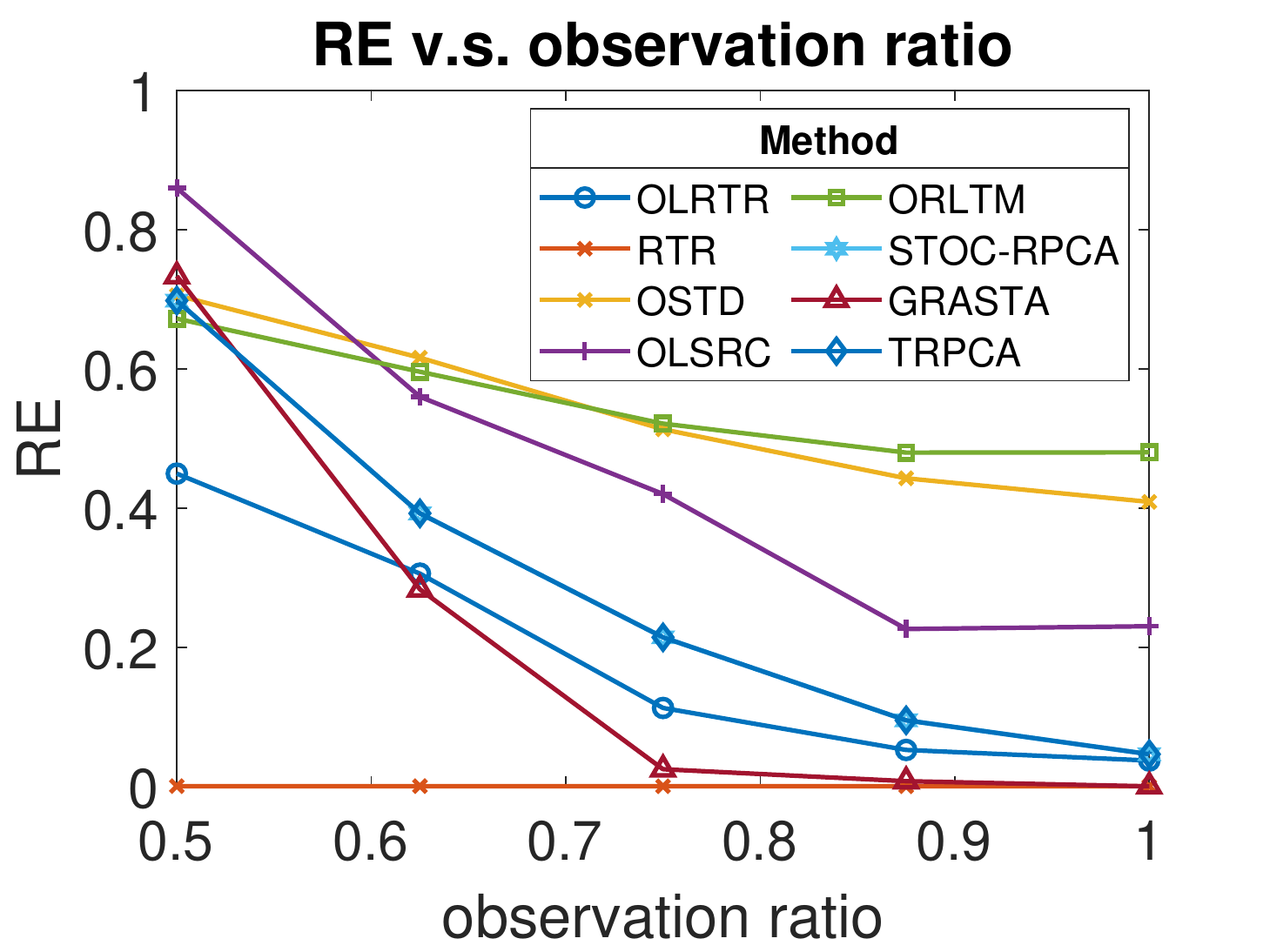}
    \caption{}
    \label{fig:adj7}
\end{subfigure}
\quad
\begin{subfigure}[t]{0.45\columnwidth}
    \includegraphics[width=\linewidth]{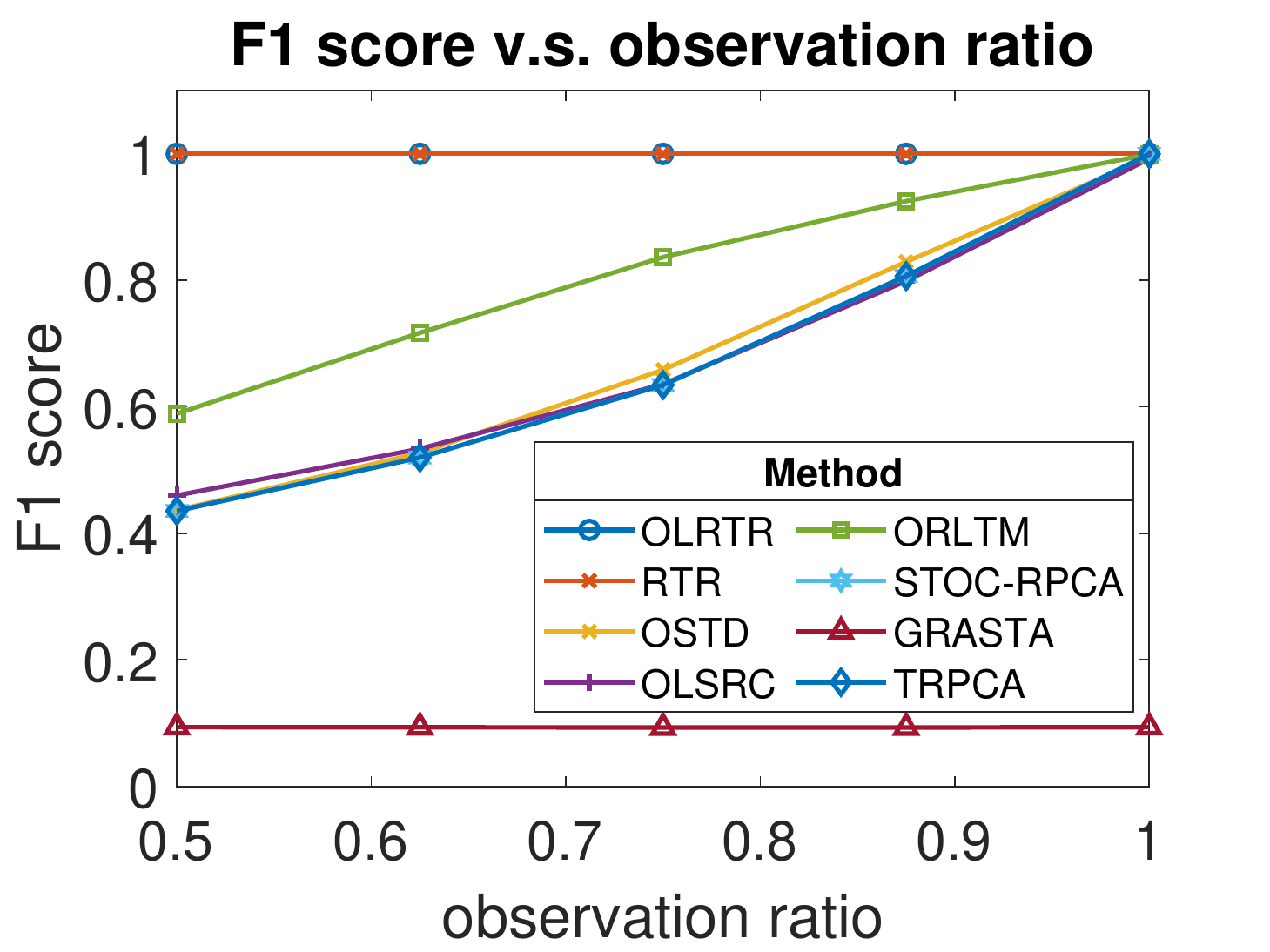}
    \caption{}
    \label{fig:adj8}
\end{subfigure}
\caption{Results of algorithms with varying observation ratio. The low-rank tensor size for each minibatch is fixed at $\mathbb{R}^{50 \times 50 \times 50}$ with a tucker rank of $(3,3,3)$, and the corruption ratio is set at 0.05. The result is an average over 5 trials.}
\label{fig:obs}
\end{figure}

Finally, we vary the observation rate. We run 100 minibatches. The low-rank tensor size for each minibatch is fixed at $\mathbb{R}^{50 \times 50 \times 50}$ with a tucker rank of $(3,3,3)$, and the corruption ratio is set at 0.05. For methods that cannot handle missing data, we linearly interpolate the missing entries, and also note that filling the missing entries with zero results in similar performance. The result is shown in Fig.~\ref{fig:obs}. We can see that the RE of all methods except RTR drop as observation ratio decreases. GRASTA deals with missing data, and we can see that its performance keeps steady for observation ratios greater than 0.75, but drops sharply as the observation ratio further decreases. Only OLRTR and RTR maintain F1 score of~1.

\begin{figure}
\centering
\includegraphics[width=0.45\linewidth]{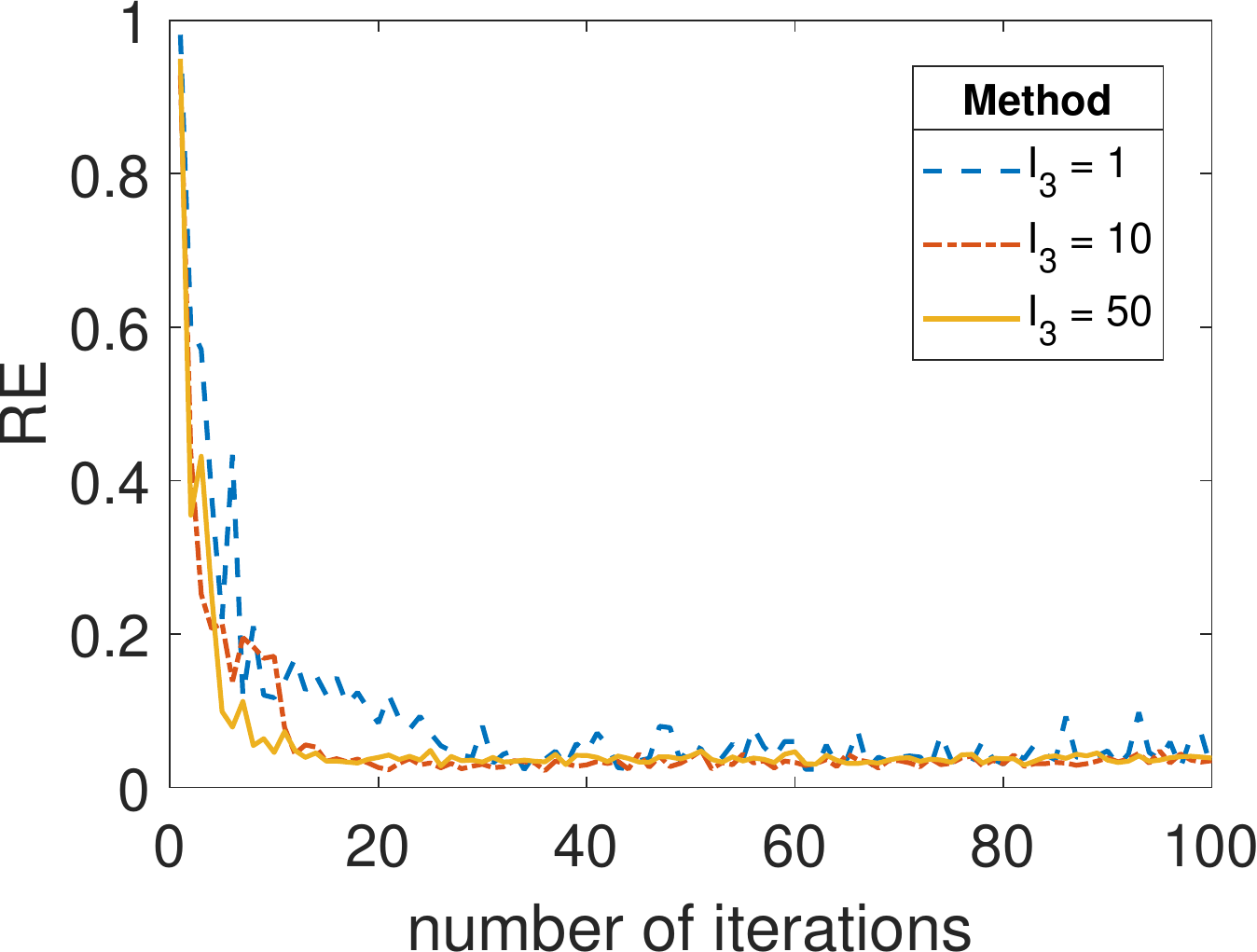}
\caption{Convergence speed of OLRTR under different minibatch size $\mathbb{R}^{50 \times 50 \times I_3}$ for $I_3 = 1,10,50$.}
\label{fig:converge}
\end{figure}

As for the convergence of OLRTR, we conduct a series of experiments with minibatch size $\mathbb{R}^{50 \times 50 \times I_3}$ for $I_3 = 1,10,50$. We fix the corruption ratio at 0.05 and the observation ratio at 0.9. The result is shown in Fig.~\ref{fig:converge}. We can see that OLRTR converges faster with larger batch size. For $I_3 = 10, 50$, OLRTR converges after 10 iterations, while for $I_3 =1$, OLRTR converges after about 30 iterations.
 
\subsection{Synthetically degraded NOAA data.} 
In this section, we apply tensor factorization on a complete NOAA dataset~\cite{noaa_data}. We use  temperature data from January to December, 2019 recorded from stationary, high-end climate sensors located at 37 USCRN monitoring sites~\cite{USCRNstation} in the US Midwest.  The accessed 37 NOAA sensors record data hourly for 24 hours a day, for 364 days, which is arranged as $\mathcal{X}\in\mathbb{R}^{37\times24\times364}$.

To test the online factorization method, we generate a synthetically degraded dataset $\mathcal{B}$ from the true temperature $\mathcal{X}_\text{true}$ that has missing data and erroneous values. We degrade the data by randomly masking 10\% of the data, and randomly modify 5\% of the tensor fibers to create outlier readings.

Given $\mathcal{B}$, we set the OLRTR hyper parameter $\alpha = 70$. To capture the daily temperature pattern, we feed in each 24-hour data $\mathcal{X}^t\in\mathbb{R}^{37\times24\times1}$ as a minibatch. For all methods, the target rank is set as $c = 20$ determined by grid search.
The results are summarized in Table~\ref{table:noaa}. We can see that among all methods, RTR has the best overall performance in RE and F1 score, whereas among all online methods, OLRTR performs the best overall. In particular, with full observation, OLRTR has comparable RE and F1 score with the batch methods, and has an F1 score at least 0.2 higher than all other online methods. Under partial observations, the RE and F1 score drop for all methods. GRASTA has slightly lower RE than OLRTR, but has a low F1 score of 0.01.  OLRTR is the only method among the online methods to have F1 score above 0.9. This experiment on NOAA data shows the validity of our methods on real world sensor networks for data recovery and anomalous sensor detection.

\begin{table}
\centering
\caption{The performance of the algorithms on NOAA data. A horizontal line separates the online methods and the batch methods (i.e., RTR \& TRPCA).}
\label{table:noaa}
            
    \sisetup{detect-weight,mode=text}
    \renewrobustcmd{\bfseries}{\fontseries{b}\selectfont}
    \renewrobustcmd{\boldmath}{}
    \newrobustcmd{\B}{\bfseries}
    \addtolength{\tabcolsep}{-4.1pt}
\begin{tabular}{l|cc|cc} 
\toprule
  \multicolumn{1}{l|}{observation rate}                      & \multicolumn{2}{c|}{1} & \multicolumn{2}{c}{0.9}  \\ 
\cmidrule{1-1} \cmidrule{2-3}\cmidrule{4-5}
metrics     & RE    & F1 score     & RE    & F1 score   \\ 
\hline
ORLTM~\cite{li2019online}    & 0.157     & 0.834     & 0.192      & 0.309      \\
STOC-RPCA~\cite{feng2013online} & 0.644  & 0.867     & 0.644     & 0.820      \\
OSTD~\cite{sobral2015online} & 0.327     & 0.294     & 0.394      & 0.179      \\
OLSRC~\cite{shen2016online}  & 0.158     & 0.800     & 0.352      & 0.199      \\
GRASTA~\cite{he2011online}   &0.085     &0.100     & \B 0.096    & 0.099      \\
OLRTR     &\B 0.053     &\B 0.974     & 0.120      &\B 0.953      \\
\hline
RTR~\cite{hu2020robust}& 0.029     &\B 0.984     &\B 0.030      & \B 0.984      \\
TRPCA~\cite{lu2019tensor}   & \B 0.012     & 0.877     & 0.078      & 0.188     \\
\bottomrule
\end{tabular}
\end{table}

\subsection{Array of Things data} 
The method is finally applied to \textit{Array of Things} (AoT), a dense urban sensor network in Chicago~\cite{Catlett:2017:ATS:3063386.3063771} that collects real-time open data on the urban environment, infrastructure, and activity. 
We construct an AoT temperature tensor as $\mathcal{X}\in\mathbb{R}^{52\times24\times183}$, representing 52 temperature sensor stations aggregated hourly, for 24 hours a day and for 365 days from March 1 2018 to March 1 2019. Approximately 16\% of the AoT data is missing in this period. We set $\alpha = 370$ in OLRTR, and the target rank at $c = 3$. 
For all online algorithms, we pass the data three epochs to refine the estimation. Due to the lack of a ground truth dataset, the recovered AOT data is quantitatively compared to the closest NOAA sensor. We use the Pearson correlation coefficient to quantify the agreement since the temperature field is spatially varying. 

\begin{figure}
\centering
\includegraphics[width=0.7\linewidth]{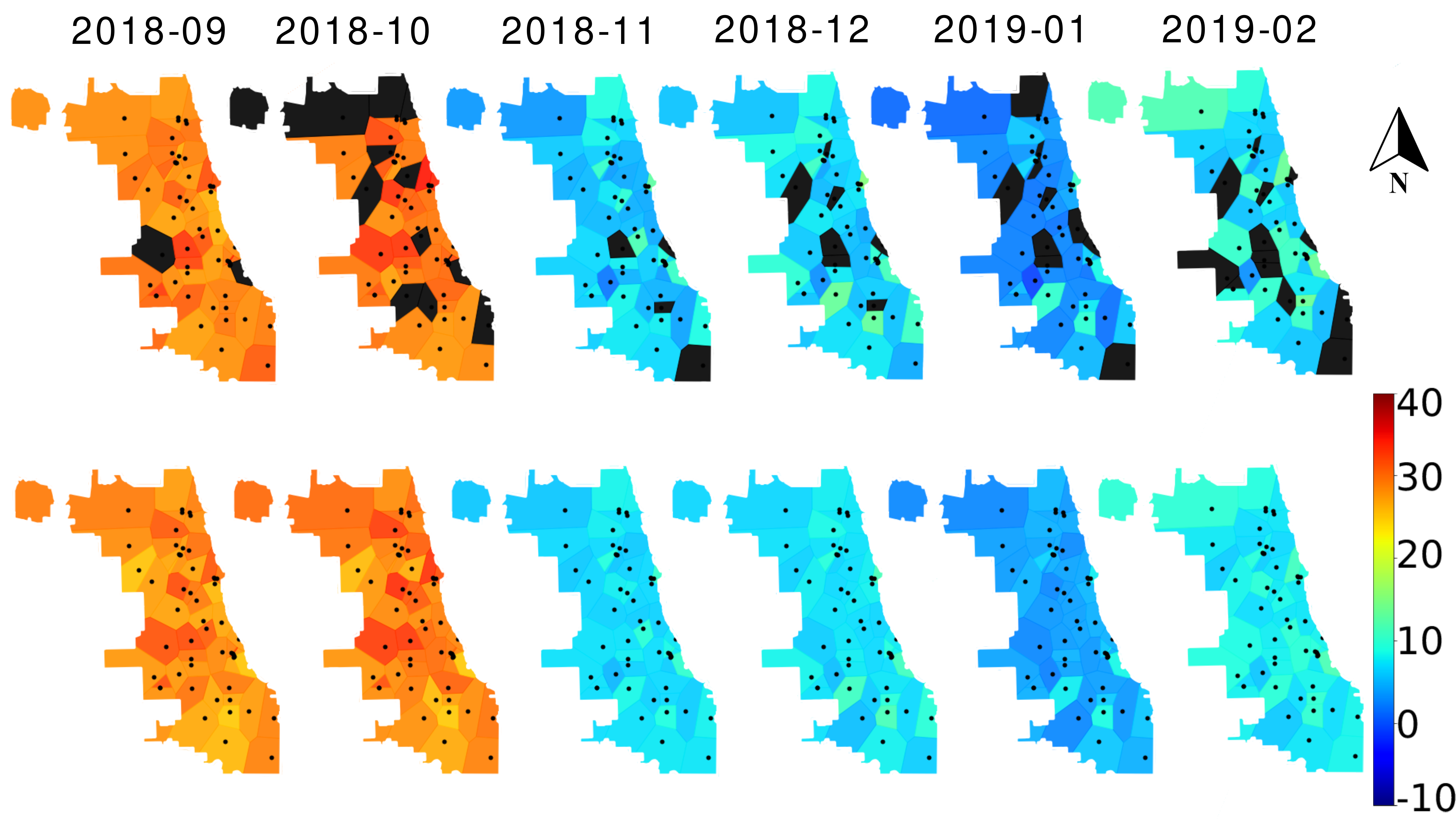}
\caption{Voronoi heat maps ($^\circ$C) at 3PM for half a year from Sept. 2018 to Feb. 2018 produced by raw (top; missing data in black) and recovered (bottom) air temperature data. Each dot marks an active AoT unit.}
\label{fig:12grid}
\end{figure}

Table~\ref{table:aot} shows the results\footnote{The GRASTA code generates warnings that the resulting matrix is singular, close to singular or badly scaled, produces NAN results, and is thus not listed.}. We see that the batch method RTR has the highest correlation of 0.98, with the same range as original input, $[-30, 43]$ Celsius. OLRTR has an comparable correlation of around 0.97, with an reasonable temperature range of $[-28, 43]$ Celsius. STOC-RPCA also has high correlation of 0.978, rightly capturing the trends, but the recovery falls in an unrealistic range of $[-1,2]$. We note that the actual record temperature of Chicago in the studied period is $[-23, 36]$, yet AOT has a larger range. This is likely due to the local conditions at the site of the sensor (e.g., lighting conditions). 
Fig.~\ref{fig:12grid} shows temperature variation in Chicago in half a year from Sept. 2018 to Feb. 2019 in the raw and recovered dataset. The method recovers a fine-grind temperature map from the raw input with outliers and missing values.

\begin{table}
\centering 
\caption{ The performance of the algorithms on AOT data, measured by the correlation coefficient between AoT and a nearby NOAA sensor, and the range of recovered temperature ($^\circ$C). We include the raw data measurements, and separate the results of batch methods (RTR, TRPCA) with the online methods into two groups.  }
\label{table:aot}   
    \sisetup{detect-weight,mode=text}
    \renewrobustcmd{\bfseries}{\fontseries{b}\selectfont}
    \renewrobustcmd{\boldmath}{}
    \newrobustcmd{\B}{\bfseries}
    \addtolength{\tabcolsep}{-4.1pt}
\begin{tabular}{c|c|ccccc|cc}
\toprule
 & Raw & STOC-RPCA  & OSTD  & OLSRC & ORLTM & OLRTR & RTR   & TRPCA \\
\hline
$r$    & 0.836         & 0.978     & 0.909         & 0.854         & 0.862         & 0.968         & \B 0.980         & 0.841         \\
Range & {[}-30, 43{]} & {[}-1, 2{]} & {[}-22, 72{]} & {[}-10, 20{]} & {[}-37, 90{]} & {[}-28, 43{]} & {[}-30, 43{]} & {[}-29, 42{]}\\
\bottomrule
\end{tabular}

\end{table}

\section{Conclusion} ~\label{Sec:conclude}
This work introduced an online tensor robust recovery method, and showed its successful application to preprossess data from large  urban sensor networks. OLRTR can detect anomalous sensors and impute missing data simultaneously, taking advantage of the multi-dimensional correlations in the dataset. Moreover, by storing and updating a small-sized dictionary that captures the underlying patterns, OLRTR can handle the data sequentially in minibatches, ensuring computational and memory efficiency in streaming systems. Extensive experiments on synthesised and real-world sensor network datasets show significant advantages of OLRTR over other established online methods, and has comparable performance with batch-based methods without the computational overhead.

While we have demonstrated the applications on temperature data, for next step we are also interested to extend the approach to accommodate other environmental sensors co-located on the AoT platform. Ultimately the cleaned data will assist its use by city planners and urban scientists interested in neighborhood-specific heat mitigation strategies to reduce adverse impacts.


\begin{acks}
This work was supported by the National Science Foundation under Grants OAC-1532133 \& CMMI-1727785, and the USDOT Eisenhower Fellowship program (No. 693JJ32045011).

\end{acks}

\bibliographystyle{unsrt}
\bibliography{ref}

\section*{Appendix}
\subsection{Online algorithm for partial observations}
The online algorithm (Algorithm~\ref{alg:OLTC}) for partial observations is derived similarly to the complete observation setting.  Starting with the batch problem~\eqref{eq:TC_batch} and relaxing the constraint as an objective function penalty, we get 
\begin{equation}
\label{eq:TC_relax}
\begin{aligned}
& \underset{\mathcal{X}_i,\mathcal{E}}{\text{min}}
& &  \frac{1}{2}\sum_{i=1}^N \|\mathcal{X_\mathnormal{i}+E+O-B}\|_F^2 + \lambda_1 \sum_{i=1}^N\|\mathbf{X}_{i(i)}\|_*+\lambda_2 \|\mathbf{E}_{(1)}\|_{2,1}\\ 
& \text{s.t.}
& & \mathcal{O}_\Omega = 0,
\end{aligned}
\end{equation}
Next, using the substitution for $\|\mathbf{X}_{i(i)}\|_*$ as in~\eqref{eq: nuclear}, we obtain:
\begin{equation}
\label{eq:TC_OL_all}
\begin{aligned}
& \underset{\mathbf{L}_i, \mathbf{R}_i, \mathcal{E}}{\text{min}}
& &  \frac{1}{2}\sum_{i=1}^N \|\mathbf{L}_i\mathbf{R}_i^T +\mathbf{E}_{(i)} + \mathbf{O}{(i)} - \mathbf{B}_{(i)}\|_F^2 +  \frac{\lambda_1}{2} \sum_{i=1}^N\left(\|\mathbf{L}_i\|^2_F + \|\mathbf{R}_i\|^2_F \right) + \lambda_2 \|\mathbf{E}_{(1)}\|_{2,1},\\
& \text{s.t.}
& & \mathcal{O}_\Omega = 0.
\end{aligned}
\end{equation}
Then, we divide the batch data into series of minibatches along the last dimension,  $\mathcal{B} =  [\mathcal{B}^1, \mathcal{B}^2, \dots, \mathcal{B}^M]$, where $\mathcal{B}^t \in \mathbb{R}^{I_1 \times I_2 \times\dots \times I_{N-1}\times I_N'}$, and $I_N' = \lfloor{I_N/M} \rfloor$. Solving~\eqref{eq:TC_OL_all} amounts to minimizing the empirical objective function:
\begin{equation}
\label{eq:TC_L}
\begin{aligned}
f_M(\mathbf{L}_i) \triangleq \frac{1}{M}\sum_{t=1}^T\sum_{i=1}^N\left[l(\mathcal{B}^t, \mathbf{L}_i) + \frac{\lambda_1}{2M}\|\mathbf{L}_i \|_F^2\right],
\end{aligned}
\end{equation}
where the loss function for each mini-batch $l(\mathcal{B}^t, \mathbf{L}_i)$ is: 
\begin{equation}
\label{eq:TC_each}
\begin{aligned}
& l(\mathcal{B}^t, \mathbf{L}_i)  =  && \underset{\mathbf{R}_i^t, \mathcal{E}^t, \mathcal{O}^t}{\text{min}}
 \frac{1}{2} \left\rVert\mathbf{L}_i\mathbf{R}_i^{t T} +\mathbf{E}_{(i)}^t + \mathbf{O}_{(i)}^t- \mathbf{B}_{(i)}^t\right\rVert_F^2 + \frac{\lambda_1}{2}  \|\mathbf{R}_i^t\|^2_F  + \lambda_2 \|\mathbf{E}_{(1)}^t\|_{2,1}\\
 & \text{s.t.}
& & \mathcal{O}_\Omega^t = 0.
\end{aligned}
\end{equation}

We take an alternative optimization approach to optimize $\mathbf{R}_i^t$, $\mathcal{E}^t$, $\mathcal{O}^t$ and $\mathbf{L}_i$. Namely, at the $t$-th time step, after accessing the new sample $\mathcal{B}^t$, we first solve for the corresponding  $\mathbf{R}_i^t$ and $\mathcal{E}^t$, $\mathcal{O}^t$, using the $\mathbf{L}_i$ obtained in last step $t-1$. Then we update $\mathbf{L}_i$, to minimize the accumulated loss given all $\{\mathbf{R}_i^\tau\}_{\tau=1}^t$ and $\{\mathcal{E}^\tau\}_{\tau=1}^t$ obtained so far. 

The update for $\mathbf{R}_i^t$ and $\mathcal{E}^t$ follows a similar approach as in~\eqref{eq:R2},~\eqref{eq:E2}. To update $\mathcal{O}^t$, we fix  $\mathbf{R}_i^t$ and $\mathcal{E}^t$ and solve:

\begin{equation}
\label{eq:O}
\begin{aligned}
 \mathcal{O}^t  &=  && \underset{\mathcal{O}}{\text{argmin}}
 \sum_{i=1}^N \left\rVert\mathbf{L}_i\mathbf{R}_i^{t T} +\mathbf{E}_{(i)}^t + \mathbf{O}_{(i)}- \mathbf{B}_{(i)}^t\right\rVert_F^2 \\
& = && \underset{\mathcal{O}}{\text{argmin}}
 \sum_{i=1}^N \left\rVert \left(\mathcal{B}^t -  \text{fold}_i(\mathbf{L}_i\mathbf{R}_i^{t T}) -\mathcal{E}^t \right) - \mathcal{O} \right\rVert_F^2  \\
 & = && \underset{\mathcal{O}}{\text{argmin}}
 N\left\rVert \frac{1}{N} \sum_{i=1}^N  \left(\mathcal{B}^t -  \text{fold}_i(\mathbf{L}_i\mathbf{R}_i^{t T}) -\mathcal{E}^t \right) - \mathcal{O} \right\rVert_F^2  \\
 & \text{s.t.}
& & \mathcal{O}_\Omega= 0,
\end{aligned}
\end{equation}
where in the second row we replace the matrix norm with its tensor norm, which are the same. For \eqref{eq:O}, we simply set $\mathcal{O}= \frac{1}{N}\sum_{i=1}^{N} \left(\mathcal{B}^t -  \text{fold}_i(\mathbf{L}_i\mathbf{R}_i^{t T}) -\mathcal{E}^t \right)$ for entries $(I_1, I_2, \dots, I_N) \in \Omega^C$, and zero otherwise. The sample update for $\mathbf{R}_i$, $\mathcal{O}$ and $\mathcal{E}$ is summarized in Algorithm~\ref{alg:R_E_O}. The stopping criterion is the same as~\eqref{eq:converge}. The update for $\mathbf{L}_i$ follows a similar approach as in the full observation case~\eqref{eq:L3}.

\begin{algorithm}
  \caption{OLRTR algorithm for partial observation}\label{alg:OLTC}
  \begin{algorithmic}[1]
     \State Given $N$-way minibatch tensors $[\mathcal{B}^1, \mathcal{B}^2, \dots, \mathcal{B}^M]$ with $\mathcal{B}^t \in \mathbb{R}^{I_1 \times I_2 \times\dots \times I_{N-1}\times I_N'}$, weighting parameters $\lambda_1, \lambda_2$, target rank $r$. Initialize dictionary $\mathbf{L}_i \in \mathbb{R}^{I_i \times r}$ and accumulation matrices $\mathbf{A}_i^t \in \mathbb{R}^{r \times r}, \mathbf{D}_i^t \in \mathbb{R}^{I_i \times r}$.
     \For{$t = 0,1,\dots , M$} 
        \State Access the $t$-th sample $\mathcal{B}^t$
        \State Solve Problem~\eqref{eq:TC_OL_all} for $\mathbf{R}_i^t$, $\mathcal{E}^t$ and $\mathcal{O}^t$ for the new sample using Algorithm~\ref{alg:R_E_O}.
        \State $\mathcal{X}^t = \frac{1}{N} \text{fold}_(i) \left(\mathbf{L}_i\mathbf{R}_i^t \right)$
        \State $A_i^t = A_i^{t-1} + \mathbf{R}_i^{t T}\mathbf{R}^t_i$;  $ \mathbf{D}_i^t = \mathbf{D}_i^{t-1} + \left(\mathbf{B}^t_{(i)} - \mathbf{E}^t_{(i)} \right)\mathbf{R}^t_i$
        \State Solve $\mathbf{L}_i^t$ with Algorithm~\ref{alg:Li}, using $\mathbf{L}_i^{t-1}$ as warm restart.
        \begin{equation*}
            \mathbf{L}_i^t = \underset{\mathbf{L}_i}{\text{argmin }} \frac{1}{2}\text{Tr}\left(\mathbf{L}_i^T \left(\mathbf{A}_i^t+\lambda_1\mathbf{I}\right)\mathbf{L}_i  \right) - \text{Tr}\left(\mathbf{L}_i^T\mathbf{D}_i^t \right)
        \end{equation*}
    \EndFor
    \State \Return low rank tensors $[\mathcal{X}^1, \mathcal{X}^2, \dots, \mathcal{X}^M]$ and outlier tensors $[\mathcal{E}^1, \mathcal{E}^2, \dots, \mathcal{E}^M]$ 
  \end{algorithmic}
\end{algorithm}

\begin{algorithm}
  \caption{Sample update for $\mathbf{R}_i$, $\mathcal{E}$ and $\mathcal{O}$}\label{alg:R_E_O}
  \begin{algorithmic}[1]
     \State Given observation $\mathcal{B}  \in \mathbb{R}^{I_1 \times\dots \times I_{N-1}\times I_N'}$, dictionary $\mathbf{L}_i  \in \mathbb{R}^{I_i \times r}$ and parameters $\lambda_1, \lambda_2$. Initialize coefficient matrix $\mathbf{R}_i \in \mathbb{R}^{I_{N \setminus i}' \times r}$, outlier tensor $\mathcal{E} \in \mathbb{R}^{I_1 \times\dots \times I_{N-1}\times I_N'}$ and compensation tensor $\mathcal{O} \in \mathbb{R}^{I_1  \times\dots \times I_{N-1}\times I_N'}$ to zero.
    \While {not converged}
        \State $\mathcal{C} \gets \left( \mathcal{B} - \mathcal{O} -\text{fold}_i(\mathbf{L}_i\mathbf{R}_i^{T})\right)$ \Comment{Update $\mathcal{E}.$ }
        \For{$j = 1,2 ,\dots, p$}
        \State  $\mathbf{E}_{(1)j} \gets  \mathbf{C}_{(1)j}\text{max}\left\{0,1-\frac{\lambda}{\mu N\|\mathbf{C}_{(1)j}\|_2}\right\}$
        \EndFor
        \For{$i = 1,2 ,\dots, N$} \Comment{Update $\mathbf{R}_i$}
        \State $\mathbf{R}_i \gets  \left(\mathbf{L}_i^T \mathbf{L}_i + \lambda_1\mathbf{I} \right)^{-1}\mathbf{L}_i^T\left( \mathbf{B}_{(i)} - \mathbf{E}_{(i)} - \mathcal{O} \right).$ 
        \EndFor
        \State $\mathcal{O} \gets  \sum_{i=1}^N \left( \mathcal{B} - \mathcal{E}  -\text{fold}_i(\mathbf{L}_i\mathbf{R}_i^{T})\right)$ \Comment{Update $\mathcal{O}.$ }
        \State  set $\mathcal{O}_\Omega $ = 0
    \EndWhile
    \State \Return $\mathbf{R}_i$, $\mathcal{E}$  and $\mathcal{O}$ 
  \end{algorithmic}
\end{algorithm}

\end{document}